# Better Prevent than Tackle: Valuing Defense in Soccer Based on Graph Neural Networks

Hyunsung Kim[1,2], Sangwoo Seo[1], Hoyoung Choi[1],
Tom Boomstra[3], Jinsung Yoon[2], and Chanyoung Park[1]

## 1. Introduction

The accurate player evaluation is critical in professional soccer, where players' market values have reached astronomical levels. However, unlike other sports where key performance indicators are more explicitly measurable, soccer lacks clear, frequent, and directly observable metrics that can fully capture a player's contribution [8]. As a result, player evaluation has traditionally relied on subjective assessment made by domain experts, who must watch numerous matches to form qualitative judgments. Despite these efforts, even high-cost player recruitments offer no guarantee of success, highlighting the need for a more quantitative approach.

With the increasing availability of event and tracking data, research on data-driven player evaluation in soccer has gained traction in recent years [6] [8] [20] [21] [23] [28] [29] [31]. Most studies have focused on valuing on-ball actions, such as passes and shots, by estimating how much each action increases a team's probability of scoring. These methods provide a more consistent and granular assessment of player contributions compared to traditional statistics.

Even with these advances, existing studies primarily focus on valuing on-ball actions, leaving a fundamental gap in crediting defensive players. Particularly, effective defending is not just about visible actions like interceptions and tackles but also about preventing dangerous offensive events before they happen [33] (As Paolo Maldini's saying goes, "If I have to make a tackle, then I have already made a mistake."). For example, if defending players effectively press the ball possessor and enforce the player to pass the ball backward by reducing the success probability of forward plays, their defensive contribution should be recognized even though they do not record any interception or tackle. Conversely, even if a play does not result in a goal, defenders should be penalized if they allow a dangerous pass or shot. Thus, valuing defensive performance is more challenging than its offensive counterpart, as it requires considering not only what defending players did with the ball, but also what they prevented (or conceded) and who should be credited (or blamed) for each.

Due to these challenges, valuing individual defensive contributions has been relatively underexplored, despite its importance in soccer analytics [13]. Bransen and Van Haaren [6] proposed Joint Defensive Impact (JDI), which compares opponents' attacking performance when a defender is on the pitch with their expectation, yet its indirect, match-level resolution cannot capture contributions in specific defensive moments. Merhej et al. [22] quantified individual defensive contributions via the potential expected threat (xT) [32] removed by successful defensive actions, but their valuation is limited to interceptions and tackles. Forcher et al. [14] and Toda et al. [34] considered more general defensive situations by modeling the probability of defensive success in the near future, but their evaluations remain at the team level rather than assigning credits to individual players. Stöckl et al. [33] analyzed which passing options defenders prevented and how threatening those options were using Graph Convolutional Networks (GCNs) [18], but their analysis is qualitative and do not propose a unified metric to compare defensive ability of multiple players. Most recently, Everett et al. [9] introduced an approach based on a Graph Attention Network (GAT) [35] that measures how removing a defender's attention weights affects the estimated pass success probability. However, they only consider passes and do not account for shots, which play a decisive role in goal-scoring. In addition, the value of a pass is defined simply as the difference of xT [32] between its start and end locations without considering contextual information.

[1] KAIST, Daejeon, South Korea
[2] Fitogether Inc., Seoul, South Korea
[3] AFC Ajax, Amsterdam, Netherlands

To address these limitations, we proposed DEFCON (**DEF**ensive **CON**tribution evaluator), a comprehensive framework that quantifies the individual contributions of defending players across every in-game situation. Using GATs [35], DEFCON estimates the success probability of each available attacking option, the expected value of that option upon its success, and the degree of responsibility each defender bears for defending it. Based on these component values, the framework computes the Expected Possession Value (EPV) at the moment of each on-ball attacking action and interprets the reduction in the attacking tam's EPV before and after their action as the defending team's contribution. DEFCON then distributes this team-level defensive value to individual defenders in a principled manner across various scenarios. When an opponent's action fails, defenders receive credit for having lowered its success probability, with additional credit assigned to the defender who directly wins the ball. When an action succeeds and increases the opponent's EPV, defenders are penalized for allowing a threatening attack. Conversely, when an action succeeds but decreases the opponent's EPV, defenders receive credit for deterring more dangerous alternatives and steering the opponent toward a less threatening choice. Through these mechanisms, DEFCON evaluates how much each defender contributes to increasing or decreasing the opponent's scoring potential in every situation, and these event-level contributions can be aggregated to quantify defensive performance over an entire match or season.

To validate our framework, we trained the component models using event and tracking data from the 2023–24 Eredivisie season and evaluated their performance on the 2024–25 season. Because no ground-truth labels exist for defensive performance, we assess DEFCON through two indirect criteria: (1) the predictive accuracy of the component models and (2) the correlation between our defensive credit and players' market values. We find that individual defenders' time-normalized credit exhibits a significant positive relationship with market value, indicating that it aligns well with domain experts' general intuition about defensive ability. Notably, the penalties imposed for allowing threatening attacks show a particularly strong correlation: defenders with smaller absolute penalties tend to have higher market values. In contrast, scores derived solely from explicit defensive actions show no positive correlation with market value, underscoring the limitations of action-based defensive evaluation and the added value of our framework. Lastly, we further showcase several practical applications, including in-game timelines of defensive contributions, spatial aggregation across pitch zones, and pairwise summaries of attacker–defender interactions. These examples demonstrate that DEFCON can effectively support scouting, player profiling, and post-match analysis in real-world professional clubs.

# 2. Framework for Defensive Valuation

The fundamental objective of defense in soccer is to reduce the opponent's probability of scoring. Accordingly, when the opponent's scoring probability decreases as a result of defensive positioning, defenders should receive positive credit; conversely, when it increases, defenders should be penalized for allowing the opponent's attack to progress. In this section, we formalize this principle by first describing how to estimate the scoring potential at each game state (Section 2.1) and how to define defensive value as the change in this potential (Section 2.2). Then, we define the defender responsibility that serves as a weighting factor for distribute this defensive value to individual defenders (Section 2.3), and introduce specific credit assignment rules under different action types and situations (Sections 2.4 and 2.5). Finally, we explain how these situation-level credits are aggregated across an entire match or season to evaluate each player's overall defensive contribution (Section 2.6). Figure 1 provides an illustrative overview of how the proposed framework computes and allocates defensive credit in a specific match situation.

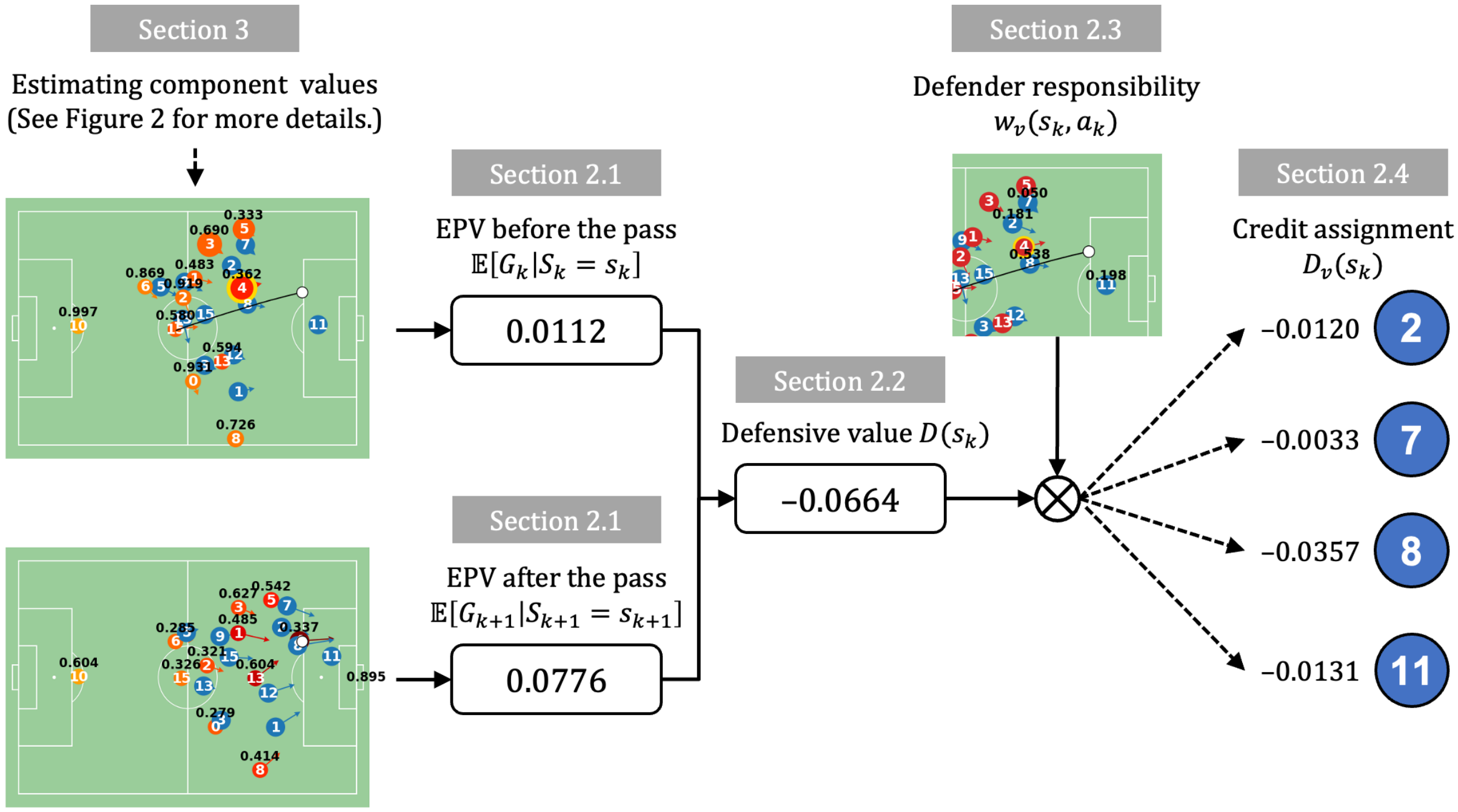

**Figure 1:** Overview of the defensive credit assignment process when conceding a threatening pass.

## 2.1. Expected Possession Value

To quantify the value of defense, we rely on the concept of *Expected Possession Value* (EPV) introduced in prior work [10] [12], which represents the scoring potential of the attacking team at a given game state. Let $s_k$ denote the state at the $k$-th on-ball attacking action and $G_k$ be the random variable representing its return, defined as 1 or $-1$ if the attacking team shortly scores/concedes a goal, and 0 otherwise. Then, EPV of state $s_k$ is defined as the expected return $\mathbb{E}[G_k|S_k = s_k]$.

Rather than directly estimating EPV using goal records as the sole supervision signal, we adopt the decomposition approach of Fernandez et al. [10] [12]. Since the ball possessor can attempt to either pass to a teammate, take on an opponent through dribbling, or shoot, we express the EPV as the combination of the probability of selecting each action $a$ and the conditional EPV if $a$ is executed:

$$\mathbb{E}[G_k|S_k = s_k] = \sum_{a \in \mathcal{A}} P(A_k = a|S_k = s_k) \cdot \mathbb{E}[G_k|S_k = s_k, A_k = a], \tag{1}$$

where $\mathcal{A} = \{a_v^{\text{pass}} | v \in \mathcal{V}^+\} \cup \{a^{\text{shot}}\}$ is the set of attacking options containing pass attempts $a_v^{\text{pass}}$ to $v \in \mathcal{V}^+$ (where $\mathcal{V}^+$ is the set of the ball possessor's teammates) and a shot $a^{\text{shot}}$. If $v$ is the ball possessor, then $a_v^{\text{pass}}$ represents dribbling attempt treated as a pass to oneself.

The *action EPV* $\mathbb{E}[G_k | S_k = s_k, A_k = a]$ given that the player selects a specific action $a$ is further decomposed by conditioning on whether the action succeeds or fails as follows:

$$\mathbb{E}[G_k | S_k = s_k, A_k = a] = \sum_{o \in \{0,1\}} P(O_k = o | s_k, a) \cdot \mathbb{E}[G_k | s_k, a, O_k = o] \qquad (2)$$

where the outcome $O_k$ of the action in $s_k$ is defined as 1 if it is successful and 0 if failed.

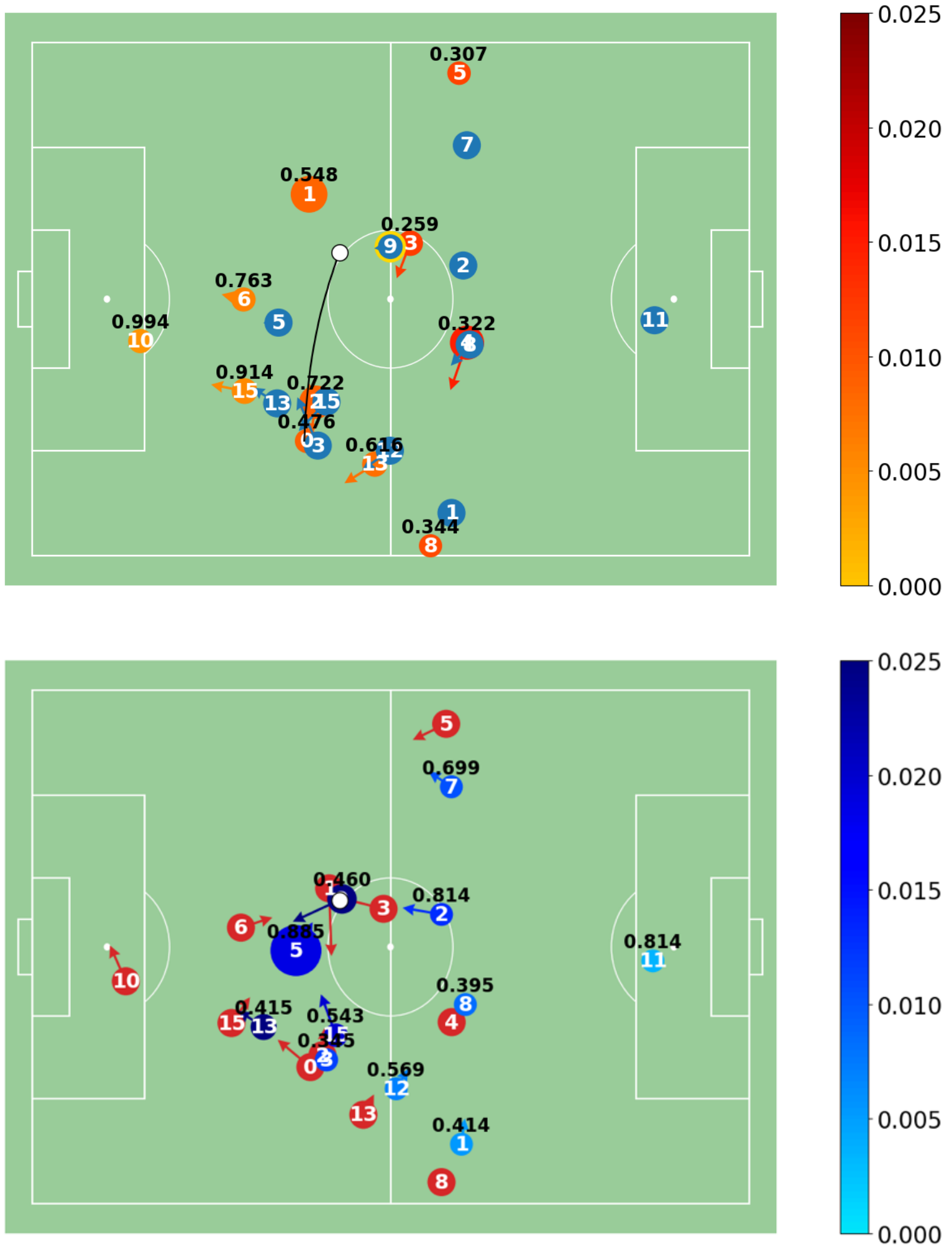

| Target | Select prob. | Success prob. | Success EPV |
|---|---|---|---|
| **Red 0** | 0.019 | 0.476 | 0.0087 |
| Red 1 | 0.312 | 0.548 | 0.0085 |
| Red 2 | 0.222 | 0.722 | 0.0087 |
| Red 3 | 0.047 | 0.259 | 0.0120 |
| Red 4 | 0.243 | 0.322 | 0.0149 |
| Red 5 | 0.010 | 0.307 | 0.0113 |
| Red 6 | 0.023 | 0.763 | 0.0059 |
| Red 8 | 0.004 | 0.344 | 0.0108 |
| Red 10 | 0.023 | 0.994 | 0.0042 |
| Red 13 | 0.044 | 0.616 | 0.0075 |
| Red 15 | 0.052 | 0.914 | 0.0052 |

| Target | Select prob. | Success prob. | Success EPV |
|---|---|---|---|
| Blue 1 | 0.000 | 0.414 | 0.0053 |
| Blue 2 | 0.000 | 0.814 | 0.0130 |
| Blue 3 | 0.006 | 0.345 | 0.0114 |
| Blue 5 | 0.789 | 0.885 | 0.0186 |
| Blue 7 | 0.001 | 0.699 | 0.0103 |
| Blue 8 | 0.000 | 0.495 | 0.0083 |
| **Blue 9** | 0.131 | 0.460 | 0.0269 |
| Blue 11 | 0.000 | 0.814 | 0.0034 |
| Blue 12 | 0.001 | 0.569 | 0.0071 |
| Blue 13 | 0.058 | 0.415 | 0.0333 |
| Blue 15 | 0.010 | 0.543 | 0.0199 |

**Figure 2:** Visualization and tabulation of component values before and after a pass. The red player 0 intended to pass to teammate 1, but the blue player 9 intercepted the pass. The size of each circle denotes the action selection probability, indicating how likely red 0 was to choose that passing option. Also, the annotation above the circle shows the pass success probability, and the color of the circle represents the EPV conditioned on the success of that pass.

In summary, EPV consists of the following three components:

(a) action selection probability $P(A_k = a | S_k = s_k)$,
(b) action success probability $P(O_k = 1 | S_k = s_k, A_k = a)$, and
(c) outcome-conditioned EPV $\mathbb{E}[G_k | S_k = s_k, A_k = a, O_k = o]$ for $o \in \{0,1\}$.

As instantiated in Figure 2, this decomposed formulation enhances interpretability by revealing which attacking options are likely to succeed and how threatening they are if successful. Moreover, it allows these components to be reused in later stages of our framework. Section 3 explains how individual components are estimated using Graph Neural Networks.

## 2.2. Team-Level Defensive Value

Once the EPV at each game state is obtained, we evaluate the defensive impact for the opponent's each on-ball action by measuring how it changes their EPV. Following Decroos et al. [8], we define the *offensive value* of the action in state $s_k$ as the change in EPV before and after the action:

$$\Delta V(s_k) = \mathbb{E}[G_{k+1} | S_{k+1} = s_{k+1}] - \mathbb{E}[G_k | S_k = s_k]. \tag{3}$$

A central principle of our framework is that defensive value should serve as the zero-sum counter part of this offensive value. Thus, the defending team's contribution for an opponent action in $s_k$ is defined as the negated offensive value:

$$D(s_k) = -\Delta V(s_k) = \mathbb{E}[G_k | S_k = s_k] - \mathbb{E}[G_{k+1} | S_{k+1} = s_{k+1}]. \tag{4}$$

For example, the red team's EPV in the left snapshot of Figure 2 is 0.0032, while the blue team's EPV in the right snapshot becomes 0.0158 after intercepting the pass. From the red team's perspective, the EPV decreases from 0.0032 to $-0.0158$, yielding an offensive value of $-0.0190$ for the pass. Conversely, the blue team receives a defensive value of $+0.0190$ for the interception.

This team-level defensive value captures how much the defending team decreases or allows to increase the opponent's scoring potential during the opponent action. The following sections detail how this defensive value is assigned to individual defenders according to the action outcome.

## 2.3. Defender Responsibility

To distribute the team-level defensive value defined in Section 2.2 to defending players, we must first quantify how responsible each of them is for defending each attacking option (i.e., a pass to each teammate or a shot). Conceptually, defenders who bear greater responsibility for defending an option should receive more credit if that option fails or is prevented, and should receive a greater penalty if the team concedes it. Therefore, we need a principled measure of such responsibility that can serve as a weighting factor for allocating the defensive value.

A pass or shot is defined as successful when the ball reaches its intended target (i.e., either a teammate or the goal). Hence, a low success probability for an action implies a high probability that one of the defenders rather than the intended target will receive the ball. In other words, a defender who is more likely to receive the ball if an action $a$ is attempted is effectively reducing the success probability of $a$ through their positioning. Therefore, if $a$ fails or is prevented, defenders with high receiver probabilities should be more credited. Conversely, they should be more penalized if $a$ succeeds, as they failed to stop $a$ despite their high defensive expectation.

Based on this intuition, we define the *defender responsibility* $w_v(s_k, a)$ of a defender $v$ for an action $a$ as being proportional to the probability $P(R_k = v | S_k = s_k, A_k = a)$ that $v$ would receive the ball if $a$ were attempted. Because the responsibility values must sum to 1 across all defenders, normalizing these probabilities yield the *failure-conditioned receiver probability* as follows:

$$w_v(s_k, a) = \frac{P(R_k = v | s_k, a)}{\sum_{u \in \mathcal{V}^-} P(R_k = u | s_k, a)} = \frac{P(R_k = v | s_k, a)}{P(O_k = 0 | s_k, a)} = P(R_k = v | s_k, a, O_k = 0), \qquad (5)$$

where $R_k$ denotes the defender who would receive the ball in $s_k$ and $\mathcal{V}^-$ is the set of players in the defending team.

This conditional receiver probability represents how likely each defender would intercept or recover the ball given the failure of an action $a$, and it serves as the basis for defensive credit assignment described in the subsequent sections. Section 3.1 details the GNN model used to estimate this responsibility and Figure 7 illustrates the resulting values assigned to players.

## 2.4. Credit Assignment for Defending Passes

When an opponent attempts a pass $a_k$ in $s_k$, the way of distributing the defensive value depends on its outcome. We consider the following five possible outcomes and elaborate on the assignment procedures case by case:

(a) the pass fails due to an on-ball defensive action (e.g., interception or tackle),
(b) the pass fails without a defensive action (e.g., the ball goes out of play or a player is offside),
(c) the pass succeeds and increases the attacking team's EPV,
(d) the pass succeeds but decreases the EPV, and
(e) the pass results in a defender's foul.

**Pass failure caused by a defensive action:** When a pass $a_k$ fails because a defender successfully performed an on-ball defensive action, the defending team receives a positive defensive value. This value must be distributed not only to the defender who performed the defensive action but also to surrounding defenders whose positioning effectively reduced the success probability of the pass.

To be specific, let $p = P(O_k = 1 | s_k, a_k)$ denote the estimated success probability of the pass. Without the defenders, the pass would almost certainly succeed, so its natural success probability would be close to 1. However, the defenders' positioning lowered this probability from 1 to $p$. The defender who intercepted the pass then eliminated the remaining success probability from $p$ to 0.

This observation motivates a decomposition of the defensive value $D(s_k)$ into two components:

- a fraction $1 - p$ of $D(s_k)$ attributable to defensive positioning that disturbed the pass,
- a fraction $p$ of $D(s_k)$ attributable to the defensive action.

The former is distributed to all defenders based on their responsibilities $w_v(s_k, a_k)$ and the latter is dedicated to the interceptor. As a result, each defender $v$ takes a defensive credit $D_v(s_k)$ as follows:

$$D_v(s_k) = \begin{cases} pD(s_k) + w_v(s_k, a_k)(1-p)D(s_k), & \text{if } v \text{ is the interceptor,} \\ w_v(s_k, a_k)(1-p)D(s_k), & \text{otherwise.} \end{cases} \qquad (6)$$

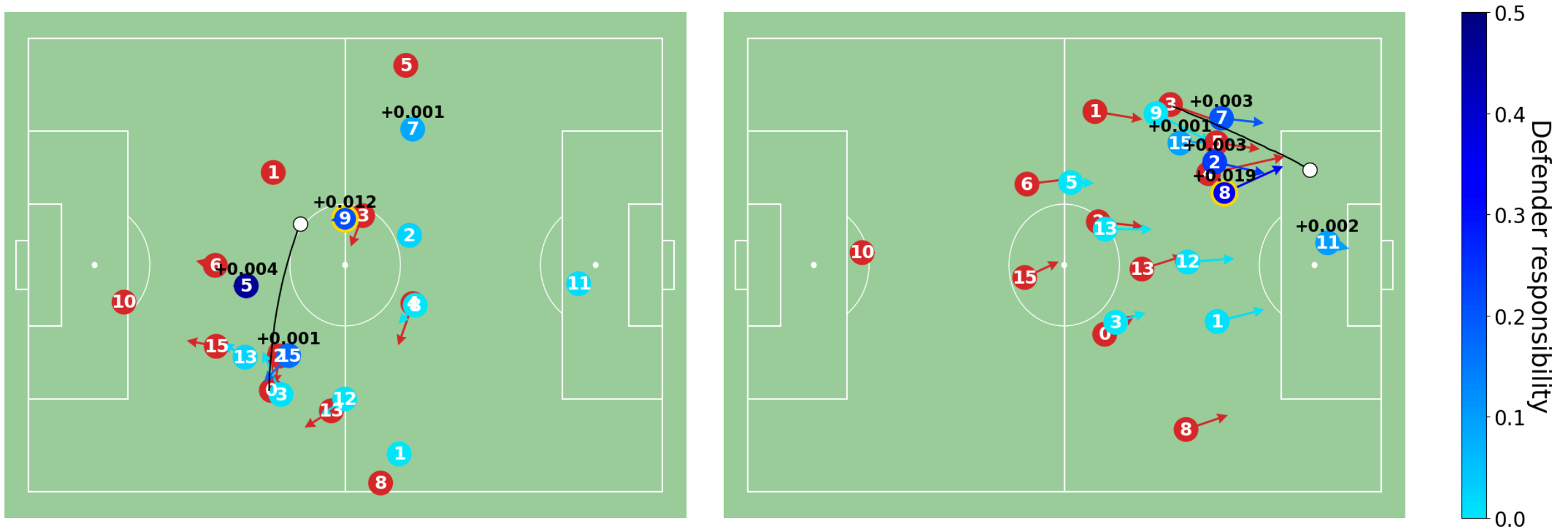

**Figure 3:** Visualization of defender responsibility and player-level defensive credit for two passes intercepted by the blue team. The color of each blue circle denotes the defender's responsibility for defending the corresponding pass, and the text above the circle indicates the defensive credit assigned to the player for contributing to the successful defense.

For example, in the left snapshot of Figure 3, the estimated success probability of the pass to red player 1 was 0.548 (see also the left snapshot in Figure 2), but the pass failed and yielded a defensive value of 0.190 as described in Section 2.2. This value was then distributed to the defending players according to Equation (6). The three players who received the largest credits are:

- blue 9 (responsibility 0.210): $0.548 \cdot 0.190 + 0.210 \cdot (1 - 0.548) \cdot 0.190 = 0.012$,
- blue 5 (responsibility 0.467): $0.467 \cdot (1 - 0.548) \cdot 0.190 = 0.004$,
- blue 15 (responsibility 0.168): $0.168 \cdot (1 - 0.548) \cdot 0.190 = 0.001$.

**Pass failure without a defensive action:** A pass may also fail without any explicit defensive action, such as when the ball goes out of play or the intended receiver is in an offside position. In such cases, the primary reason for the failed pass is the defending players' positioning, which reduced the pass success probability and disturbed the ball possessor from completing the pass. Thus, the entire defensive value $D(s_k)$ is attributed to defenders proportionally to their responsibilities for defending the pass:

$$D_v(s_k) = w_v(s_k, a_k) D_v(s_k) \tag{7}$$

**Successful pass leading to an increase in EPV:** When a pass $a_k$ is successful and results in an increase in the attacking team's EPV, a penalty is imposed to the defending team for allowing the ball to move into a more threatening location. In this case, defenders with higher responsibility values had a relatively high probability of intercepting the ball, but the pass eventually succeeded despite their high defensive expectation. Therefore, they should take a larger share of the penalty. Accordingly, like the previous case, we allocate the defensive penalty $D(s_k)$ (which is negative in this case) to defenders proportionally to their responsibilities, as illustrated in Figure 4:

$$D_v(s_k) = w_v(s_k, a_k) D_v(s_k) \tag{8}$$

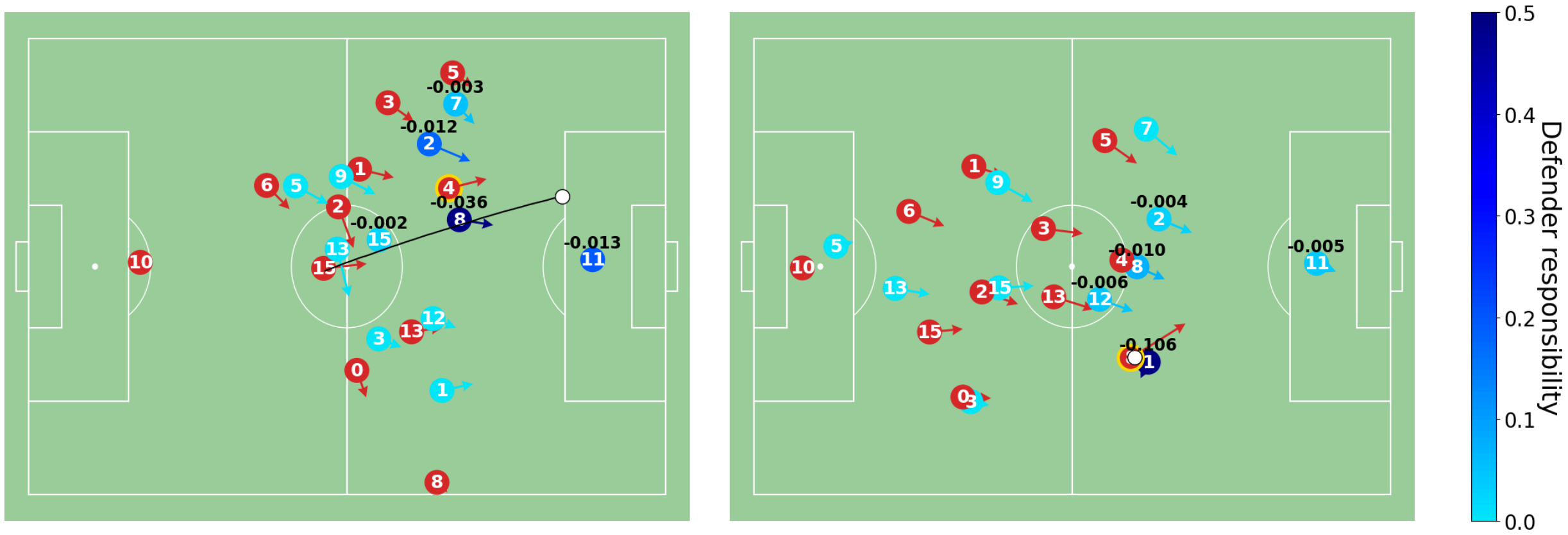

**Figure 4:** Visualization of defender responsibility and player-level defensive credit for a successful pass and a successful dribble by the red team. The color of each blue circle denotes the defender's responsibility for defending the corresponding pass, and the text above the circle indicates the penalty imposed to the player for conceding the action.

**Successful pass leading to a decrease in EPV:** When a pass $a_k$ is completed but results in a decrease in the attacking team's EPV, the pass itself is typically not a preferred choice for the attacking team. Rather, the selection of $a_k$ reflects the defending team's effective positioning, which has neutralized more threatening alternatives and forced the attacker to select a less valuable option. In such situations, the defending team deserves the positive defensive value $D(s_k)$ for deterring those higher-value options. Because multiple threatening options may exist, this defensive value should then be allocated to individual defenders according to both the relative importance of deterring each option and the defender responsibilities for the option.

Specifically, we define the set of threatening options in the given state $s_k$ as those that would increase the EPV if successful:

$$\mathcal{A}_k^+ = \{a \in \mathcal{A} \mid \mathbb{E}[G_k | s_k, a, O_k = 1] > \mathbb{E}[G_k | s_k]\}. \tag{9}$$

Before assigning defender-level credits, we divide the team-level defensive value $D(s_k)$ across the threatening options $a \in \mathcal{A}_k^+$ according to their relative importance. Intuitively, $a$ receives a higher share $D(s_k, a)$ when (i) its successful completion would substantially raise the EPV, and (ii) defenders have significantly reduced its success probability, resulting in a much lower action EPV $\mathbb{E}[G_k | s_k, a]$. That is, we assign a defensive value $D(s_k, a)$ for deterring each threatening option $a \in \mathcal{A}_k^+$ proportionally to the potential increase in the EPV conditioned its success:

$$D(s_k, a) \propto \mathbb{E}[G_k | s_k, a, O_k = 1] - \mathbb{E}[G_k | s_k, a], \tag{10}$$

and assign $D(s_k, a) = 0$ for non-threatening options $a \notin \mathcal{A}_k^+$.

Once we obtain the defensive value $D(s_k, a)$ for deterring each option $a$, we distribute it to individual defenders. As discussed in Section 2.3, we use the defender responsibility $w_v(s_k, a)$ as a weighting factor to compute the contribution $D_v(s_k, a)$ of defender $v$ to deterring $a$ as follows:

$$D_v(s_k, a) = w_v(s_k, a) D(s_k, a). \tag{11}$$

Consequently, aggregating $D_v(s_k, a)$ across the threatening options $a \in \mathcal{A}_k^+$ yields the overall defensive credit $D_v(s_k)$ assigned to $v$ in the given situation $s_k$:

$$D_v(s_k) = \sum_{a \in \mathcal{A}_k^+} D_v(s_k, a) = \sum_{a \in \mathcal{A}_k^+} w_v(s_k, a) D(s_k, a). \tag{12}$$

**Foul committed by a defender:** When the defending team commits a foul, the next on-ball attacking action becomes a free kick or a penalty kick granted to the attacking team. In such cases, the defensive value $D(s_k)$ is the change in EPV between the current state and the set-piece state that immediately follows the foul. Because the foul results from the action of a single defender, the entire defensive penalty is exclusively assigned to the player who committed the foul, i.e., $D_v(s_k) = D(s_k)$. For example, since we uniformly set the EPV for all penalty kicks to 0.7884, a player who commits a foul resulting in a penalty kick is penalized by $\mathbb{E}[G_k | s_k] - 0.7884$.

## 2.5. Credit Assignment for Defending Shots

As in the case of passes descried in Section 2.4, the allocation of defensive value for shots depends on the outcome of the attempt. Unlike the conventional definition that regards a shot as successful only when it results in a goal, we define the success of a shot as not being blocked by an outfield player. This definition reflects the fundamentally different defensive roles of outfield players and goalkeepers: the former primarily aims to block shots, whereas the latter is responsible for saving unblocked shots in front of the goal. Accordingly, we treat the defensive effort of blocking a shot in parallel with the interception of a pass, and distinguish two possible shooting outcomes: the shot is blocked or not blocked by an outfield player.

**Blocked shot:** A blocked shot is treated in the same manner as a pass intercepted by a defender, as both situations involve a defender stopping the ball from reaching its intended target through direct intervention in the ball's trajectory. Therefore, the defensive value $D(s_k)$ is distributed following the same principle as in Equation (6), with the only difference being that $p = P(O_k = 1 | s_k, a_k)$ now denotes the probability that the shot $a_k = a^{\text{shot}}$ is not blocked.

**Unblocked shot:** When a shot is not blocked by an outfield player, the attacking team's EPV diverges significantly depending on whether the shot results in a goal. If the attacking team scores, their EPV rises sharply to 1. Otherwise, it typically decreases, except in situations where the attacking team immediately obtains a rebound shooting opportunity.

Regardless of this divergence, failing to block the shot already constitutes a defensive fault from the outfield defenders' perspective. That is, it means that they have allowed the ball to reach the goal area, thereby failing in their primary defensive duty. Thus, they should receive a penalty independent of whether the shot leads to a goal.

To quantify this penalty, we compare the attacking team's current EPV $\mathbb{E}[G_k | s_k]$ with the *unblocked-shot expected goal* (UxG) defined as

$$U(s_k) = P(G_k = 1 | s_k, a^{\text{shot}}, O_k = 1), \tag{13}$$

which represents the scoring probability if a shot would not be blocked. The difference between these two values (which is typically negative) is distributed among the outfield defenders in proportion to their responsibilities $w_v(s_k, a^{\text{shot}})$:

$$D_v(s_k) = w_v(s_k, a^{\text{shot}})\big(\mathbb{E}[G_k|s_k] - U(s_k)\big). \tag{14}$$

Because these responsibilities reflect the likelihood that each defender would have been the blocker if the shot had been blocked, the goalkeeper receives zero responsibility weight.

The goalkeeper's credit, in contrast, depends on the consequence of the unblocked shot. If the shot is off target and requires no goalkeeper intervention, the goalkeeper $v_{\text{GK}}$ receives zero credit. If the shot is on target, $v_{\text{GK}}$ receives $U(s_k) - \mathbb{E}[G_{k+1}|s_{k+1}]$, which may be positive or negative depending on the subsequent event. Specifically, if $v_{\text{GK}}$ saves the shot and transitions the game into a state with a smaller $\mathbb{E}[G_{k+1}|s_{k+1}]$, they obtain positive credit for the save. Conversely, if the save leads to a more dangerous rebound opportunity, then $\mathbb{E}[G_{k+1}|s_{k+1}]$ may exceed $U(s_k)$, yielding a negative credit (penalty). If the shot results in a goal, then $\mathbb{E}[G_{k+1}|s_{k+1}] = 1$, imposing the goalkeeper a large penalty of $U(s_k) - 1$. In summary, the credit for the goalkeeper is

$$D_{v_{\text{GK}}}(s_k) = \begin{cases} U(s_k) - \mathbb{E}[G_{k+1}|s_{k+1}], & \text{if the shot is on target,} \\ 0, & \text{if the shot is off target.} \end{cases} \tag{15}$$

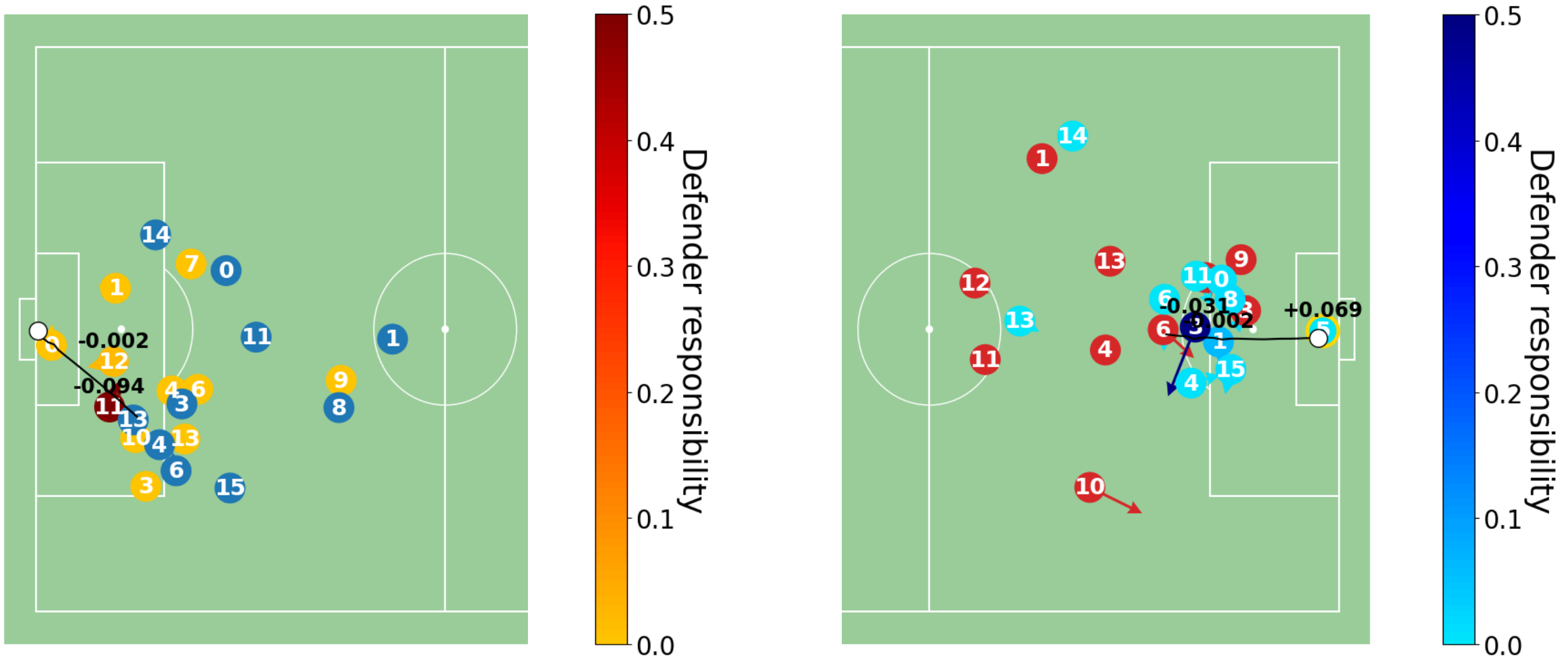

**Figure 5:** Visualization of defensive credit assignment in unblocked-shot situations. The left and right snapshots show examples of a shot off target and a saved shot on target, respectively. Circle colors for the defending team represent each defender's responsibility for defending the shot, and the text above each circle indicates the defensive credit or penalty.

Figure 5 instantiates how defensive credit is assigned in unblocked-shot situations. In the left snapshot, the attacking team's EPV before the shot was 0.024, while its UxG was 0.121. This means that the defenders allowed the situation to reach a state in which the scoring probability was 0.121, so they share a penalty of $0.024 - 0.121 = -0.097$ according to their individual responsibilities.

In the right snapshot, the EPV before the shot was 0.035 and the UxG was 0.069. As in the previous case, the defenders incur a penalty of $0.035 - 0.069 = -0.034$ for failing to block the shot.

However, in this example the shot was on target and the goalkeeper saved it. Therefore, the goalkeeper receives positive credit of 0.069 for reducing the scoring probability from 0.069 to 0.

Through this formulation, our framework assigns meaningful penalties to defenders even for non-goal shots, which occur far more frequent than shots resulting in goals. This enables us to better quantify the defender's ability to prevent dangerous shooting opportunities before they develop.

## 2.6. Aggregating Defensive Contributions Across Matches

The situation-level defensive credits defined in the previous subsections can be aggregated over an entire match or season to evaluate each player's overall defensive contribution. For interpretability, we group these credits into four categories:

(a) Intercept: credit for directly winning the ball through on-ball defensive actions,
(b) Disturb: credit for inducing the failure of the opponent's attack by lowering its success probability through effective positioning,
(c) Deter: credit for deterring more threatening attacking options when the opponent completes a less valuable action that reduces their EPV, and
(d) Concede: penalty for allowing threatening attacks that increase the opponent's EPV.

Figure 6 shows the aggregated credits for defenders in a match where the home team defeated the away team 5–0. Because the home team dominated possession, the away defenders were involved in many on-ball defensive actions such as interceptions and tackles and consequently accumulated large on-ball credits (blue bars). However, they also conceded many threatening attacks, leading to far larger penalties (red bars), which ultimately produced substantially lower net credits (purple bars) compared to the home defenders.

This example underscores a key limitation of existing evaluating metrics that rely solely on on-ball defensive actions. That is, they can misleadingly suggest that the away defenders performed better simply because they executed more on-ball defensive actions. In contrast, our framework appropriately penalizes defenders for allowing dangerous situations, yielding a more reliable and comprehensive assessment of defensive performance.

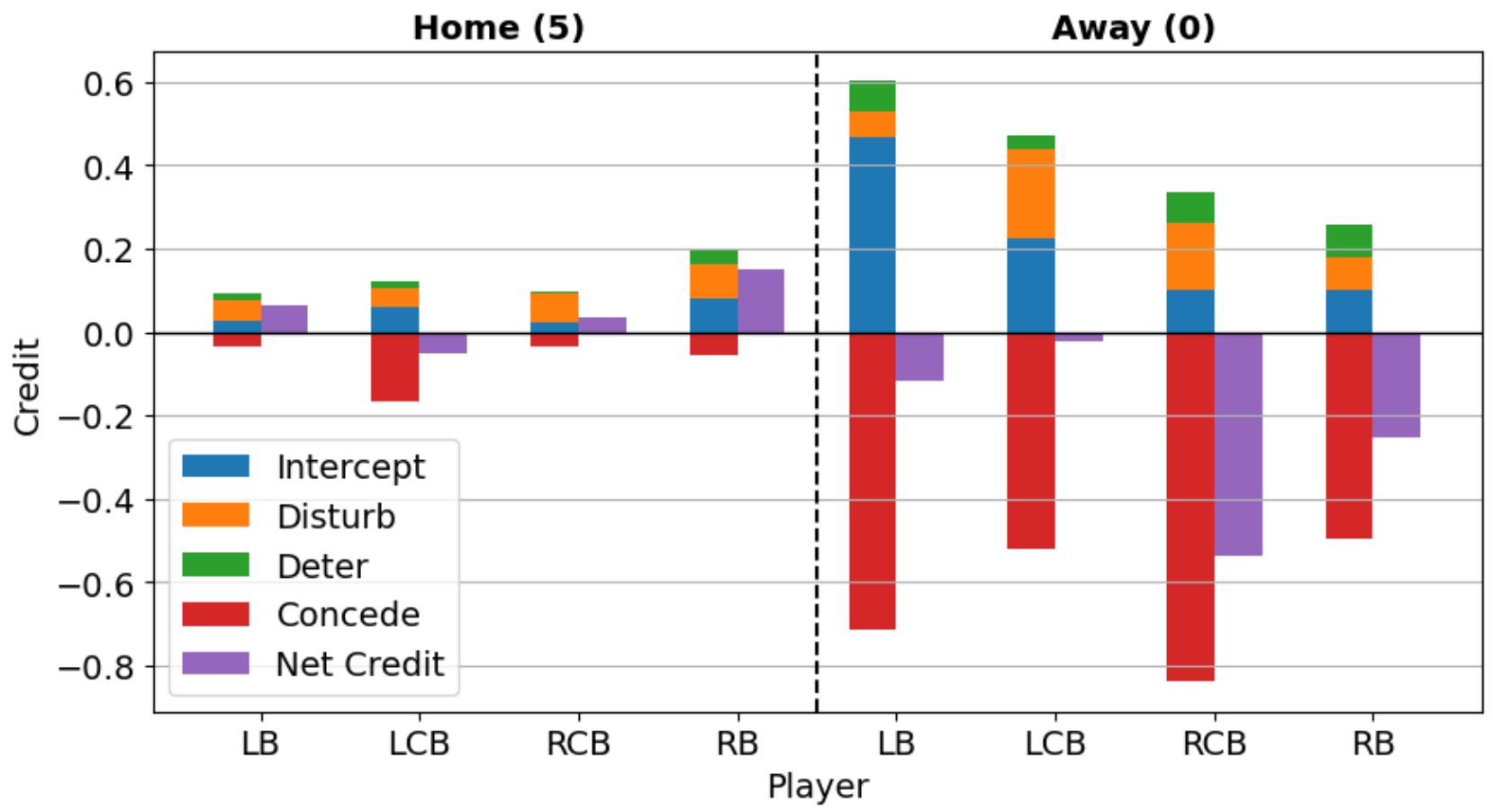

**Figure 6:** Category-wise and net defensive credits aggregated per 90 minutes for each defender in a given match. The numbers at the top indicate the final score, where the home team defeated the away team 5–0.

# 3. Estimation of Component Values

To implement the credit assignment framework introduced in Section 2, we must estimate the following underlying component values:

(a) action selection probability $P(A_k = a|S_k = s_k)$ (Section 2.1),
(b) action success probability $P(O_k = 1|S_k = s_k, A_k = a)$ (Section 2.1),
(c) outcome-conditioned EPV $\mathbb{E}[G_k|S_k = s_k, A_k = a, O_k = o]$ for $o \in \{0,1\}$ (Section 2.1), and
(d) defender responsibility $w_v(s_k, a) = P(R_k = v|S_k = s_k, A_k = a, O_k = 0)$ (Section 2.3).

Because player movement patterns differ substantially between passes and shots, we estimate the action success probability separately for the two action types, namely, the *pass success probability* $P\left(O_k = 1 \middle| s_k, a_v^{\text{pass}}\right)$ and the *shot success probability* $P\left(O_k = 1 \middle| s_k, a^{\text{shot}}\right)$. As we define the success of a shot as not being blocked by an outfield player, we calculate the shot success probability by estimating the *shot blocking probability* $P\left(O_k = 0 \middle| s_k, a^{\text{shot}}\right)$ and subtracting it from 1.

Outcome-conditioned EPV also requires separate modeling for passes and shots. For passes, we follow existing approaches [8] [31] that decomposes the conditional EPV into *goal-scoring* and *goal-conceding probabilities* and estimate them independently:

$$\mathbb{E}\left[G_k \middle| s_k, a_v^{\text{pass}}, O_k = o\right] = P\left(G_k = 1 \middle| s_k, a_v^{\text{pass}}, O_k = o\right) - P\left(G_k = -1 \middle| s_k, a_v^{\text{pass}}, O_k = o\right). \quad (16)$$

For shots, we treat the EPV for a blocked shot as zero, while for unblocked shots we adopt the *unblocked-shot expected goal* $U(s_k) = P\left(G_k = 1 \middle| s_k, a^{\text{shot}}, O_k = 1\right)$ introduced in Section 2.5:

$$\mathbb{E}\left[G_k \middle| s_k, a^{\text{shot}}, O_k = 1\right] = U(s_k), \qquad \mathbb{E}\left[G_k \middle| s_k, a^{\text{shot}}, O_k = 0\right] = 0. \quad (17)$$

In summary, our framework trains separate predictive models for the following component values:

(a1) action selection probability $P(A_k = a|s_k)$,
(b1) pass success probability $P\left(O_k = 1 \middle| s_k, a_v^{\text{pass}}\right)$,
(b2) shot blocking probability $P\left(O_k = 0 \middle| s_k, a^{\text{shot}}\right)$,
(c1) outcome-conditioned goal-scoring probability $P\left(G_k = 1 \middle| s_k, a_v^{\text{pass}}, O_k = o\right)$,
(c2) outcome-conditioned goal-conceding probability $P\left(G_k = -1 \middle| s_k, a_v^{\text{pass}}, O_k = o\right)$,
(c3) unblocked-shot expected goal (UxG) $U(s_k) = P\left(G_k = 1 \middle| s_k, a^{\text{shot}}, O_k = 1\right)$, and
(d1) defender responsibility $w_v(s_k, a) = P(R_k = v|s_k, a, O_k = 0)$.

Except for UxG (c3), we employ Graph Neural Networks (GNNs) as in previous studies [3] [9] [27] [30] [33] [36] [37] to effectively model complex interactions among multiple players. In contrast, UxG is modeled using a logistic regression mainly based on shot location to leverage publicly available event data [24] that contains much more shots than the tracking dataset used in this study. The following subsections describe the training procedures for these tasks in detail.

## 3.1. Graph Neural Network Models

For each on-ball attacking action, we construct a fully connected graph $\mathcal{G} = (\mathcal{V}, \mathcal{E})$ in which all players on the pitch and two goals are represented as nodes $v \in \mathcal{V}$, with every pair of nodes connected by an edge $e \in \mathcal{E}$. Inspired by earlier studies [2] [11] [19] [29] [30] [31] on contextual analysis using tracking data in soccer, we represent each node with 25 node features as follows:

- four binary attributes: indicators for whether the node corresponds to the ball carrier, a teammate of the ball carrier, a goalkeeper, and a goal.
- six running features: the node's $(x, y)$ location, $(x, y)$ velocity, speed, and acceleration.
- three relative features to the goal: the distance to the defending team's goal and the sine and cosine of the angle between the node-goal line and the x-axis (i.e., the lower sideline).
- six ball-related features: the height of the ball (shared across all nodes), the distance to the ball, the sine and cosine of the angle between the node-ball line and the x-axis, and the sine and cosine of the angle between the node's velocity and the ball carrier's velocity.
- four opponent-context features: the distance to the nearest opponent, the number of opponents within a 3m radius of the node, the number of opponents closer to their goal than the node, and the number of opponents inside the triangular region formed by the node and the opposing team's two goalposts.
- two passing-line features: the distance from the potential passing line to its nearest opponent and the number of opponents in a 10m-wide corridor around the passing line, where the passing line is the line segment between the ball possessor and the node.

**Table 1:** Summary of predictive tasks for estimating the component values, including how each task is formulated within the GNN architecture, which action types and labels are used for training.

| Task | Output activation | Output size | Pass | Dribble | Shot | Supervision label |
|---|---|---|---|---|---|---|
| (a1) Action selection | Softmax over teammate nodes | 1 per node | ✓ | ✓ | ✓ | Actual action taken |
| (b1) Pass success | Node-wise sigmoid | 1 per node | ✓ | ✓ | | Whether the selected pass succeeded |
| (b2) Shot blocking | Average pooling and sigmoid | 1 per graph | | | ✓ | Whether the shot was blocked |
| (c1) Goal-scoring | Node-wise sigmoid | 2 per node | ✓ | ✓ | | Whether the actual action and outcome led to shortly scoring/conceding a goal |
| (c2) Goal-conceding | | | ✓ | ✓ | | |
| (d1) Responsibility | Softmax over opponent nodes | 1 per node | ✓ (failed actions) | | | Actual receiver of the action |

We additionally include two edge features: for each pair of nodes: the Euclidean distance between them and a binary indicator for whether they belong to the same team.

Because each component value requires a distinct prediction target, we train a separate Graph Attention Network (GAT) model [35] for the task. All models take the above graph $\mathcal{G}$ represented by the node features $\mathbf{X} \in \mathbb{R}^{|\mathcal{V}| \times 25}$ and the edge features $\mathbf{E} \in \mathbb{R}^{|\mathcal{E}| \times 2}$ as input and produce node embeddings $\mathbf{H} = (\mathbf{h}_v)_{v \in \mathcal{V}}$ by two layers of GAT convolution:

$$\mathbf{H} = \mathrm{GAT}_{\theta_2}\left(\mathrm{GAT}_{\theta_1}(\mathbf{X}, \mathbf{E})\right), \tag{18}$$

where their output structures and supervision labels differ by task. Table 1 summarizes these differences, and the following paragraphs describe each model in more detail.

**(a1) Action selection:** This task aims to predict the node to which the ball possessor will attempt to send the ball. Since the ball possessor does not intend to pass to an opponent, we apply node-wise MLPs $\phi$ and a softmax only to the node embeddings $\mathbf{H}^+ = (\mathbf{h}_v)_{v \in \mathcal{V}^+}$ of the ball possessor's teammates and the target goal:

$$\hat{\mathbf{y}}^+ = (\hat{y}_v)_{v \in \mathcal{V}^+} = \text{softmax}\left(\text{MLP}_\phi(\mathbf{H}^+)\right). \tag{19}$$

The output values $\hat{\mathbf{y}}^+ = (\hat{y}_v)_{v \in \mathcal{V}^+}$ represent the probabilities of passing (or shooting) to the candidate nodes, satisfying $\sum_{v \in \mathcal{V}^+} \hat{y}_v = 1$. The model is trained using cross-entropy loss against the one-hot vector of the true attempt.

To correctly supervise the model, we need to identify the intended receivers of failed passes, since their observed receivers are opponents. Following the approach of previous studies [25] [31], we treat the teammate closest to the endpoint of a failed pass $a$ in terms of distance and angle as its intended receiver. That is, we find the teammate $v$ that maximizes the following score:

$$r(a, v) = \frac{\min_{u \in \mathcal{V}^+} \text{Dist}(a, u)}{\text{Dist}(a, v)} \cdot \frac{\min_{u \in \mathcal{V}^+} \text{Angle}(a, u)}{\text{Angle}(a, v)}, \tag{20}$$

where $\text{Dist}(a, v)$ denotes the distance between the endpoint of $a$ and the location of the teammate $v$ at the moment of reception, and $\text{Angle}(a, v)$ is the angle between the passing line of $a$ and the line connecting the passer and $v$.

**(b1) Pass success:** We independently apply a node-wise MLP $\phi$ and a sigmoid activation $\sigma$ to each node embedding to produce the probability that the associated action would succeed if attempted:

$$\hat{y}_v = \sigma\left(\text{MLP}_\phi(\mathbf{h}_v)\right), \quad v \in \mathcal{V}. \tag{21}$$

During training, we focus only on the probability $\hat{y}_{v_k} \in \mathbb{R}$ for the observed pass $v_k$, and the model is supervised using binary cross-entropy against whether that action succeeded or failed.

**(b2) Shot blocking:** Unlike the previous tasks, predicting whether a shot attempt in the given state will be blocked is modeled as a graph classification task. Instead of producing node-wise logits, the GAT aggregates node embeddings through average pooling to obtain a single graph embedding $\mathbf{h}_\mathcal{G}$. Then, it estimates a shot blocking probability $\hat{y} \in \mathbb{R}$ through an MLP $\psi$ with a sigmoid activation $\sigma$:

$$\hat{y} = \sigma\left(\text{MLP}_\psi(\mathbf{h}_\mathcal{G})\right) = \sigma\left(\text{MLP}_\psi\left(\frac{1}{|\mathcal{V}|}\sum_{v \in \mathcal{V}} \mathbf{h}_v\right)\right). \tag{22}$$

The model is trained using binary cross-entropy against whether the shot is actually blocked.

A key challenge in this task is that players rarely attempt shots when they expect them to be blocked, so training only on observed shots underestimate the true blocking probability. To address this issue and the inherent scarcity of shots in the dataset, we augment the training data by treating certain passes as proxy failed shots. Specifically, we include passes occurring in regions with UxG greater than 0.05 and with a defender positioned between the ball possessor and the goal.

**(c1 & c2) Outcome-conditioned goal-scoring/conceding:** Estimating the outcome-conditioned goal-scoring probability is formulated as a node classification task in which each node $v \in \mathcal{V}$ produces two probabilities $(\hat{y}_v^+, \hat{y}_v^-)$: one representing the probability that the team will shortly scores a goal if the pass to $v$ succeeds, and the other corresponding to the probability if it fails.

$$(\hat{y}_v^+, \hat{y}_v^-) = \sigma\left(\mathrm{MLP}_\phi(\mathbf{h}_v)\right), \quad v \in \mathcal{V}. \tag{23}$$

During training, we focus only on the node $v_k$ corresponding to the executed action and select one of its two probabilities associated with its observed outcome. The selected probability is then compared using binary cross-entropy against the ground-truth indicator of whether the attacking team scores within the next ten events [8] [31]. A parallel procedure is used to estimate the outcome-conditioned goal-conceding probability, with the ground truth indicating whether the current attacking team concedes a goal within the next ten events.

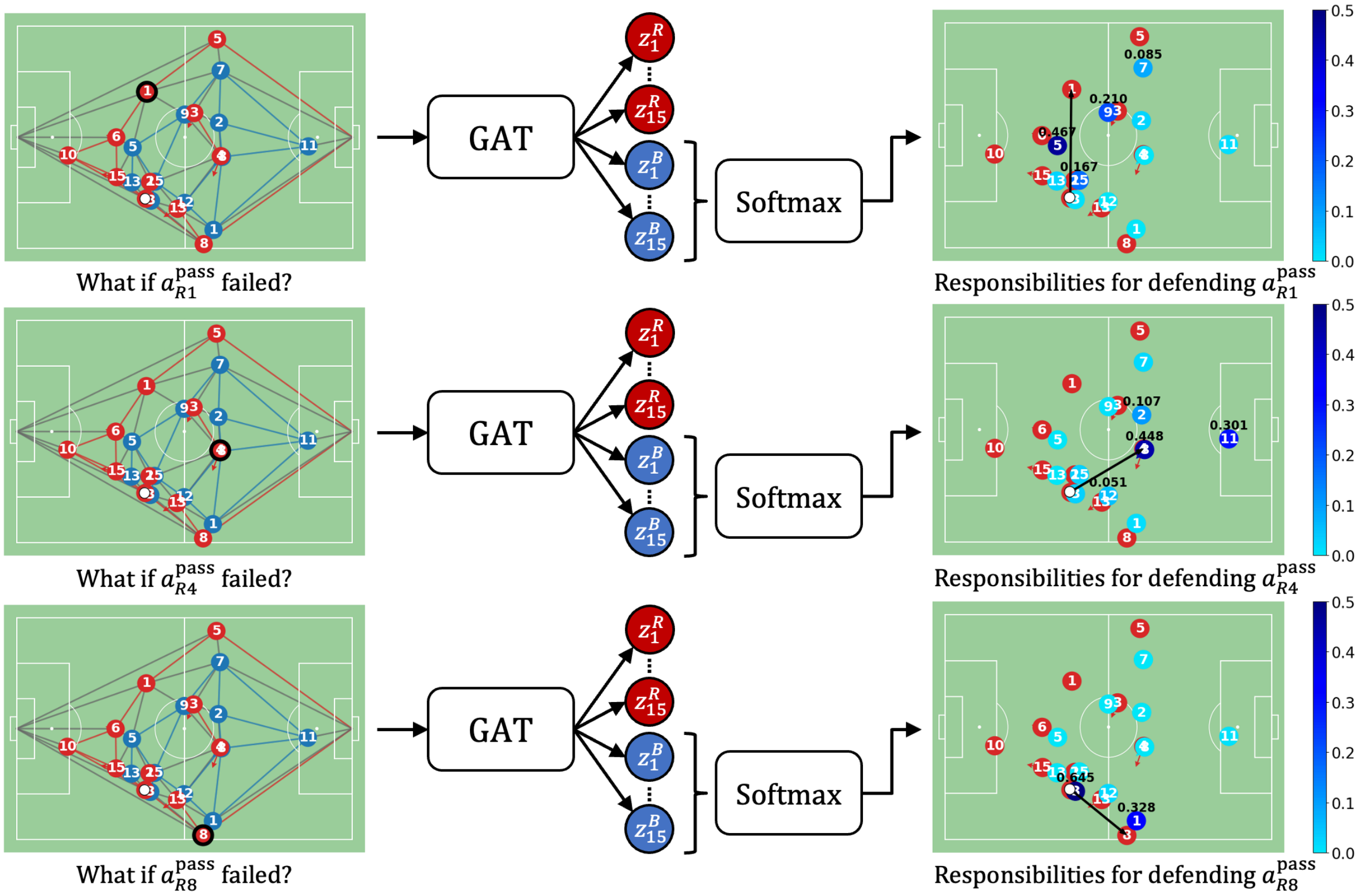

**Figure 7:** Illustration of the defender responsibility estimation model. Each left panel shows an input graph conditioned on a hypothetical pass. A binary indicator is added to the node features, where only the highlighted node is assigned a value of 1 and all others are assigned 0. The right panels display the corresponding model outputs, where the blue-team nodes are colored and annotated according to their predicted responsibilities for defending the pass to the highlighted node.

**(d1) Defender responsibility:** We model the task of predicting which defending player would receive the ball if a given pass or shot were to fail. This corresponds to selecting one receiver node conditioned on the specific action being attempted and its failure, as illustrated in Figure 7. To encode this conditioning, we extend the original 25 node features with an additional binary indicator specifying the action of interest, assigning 1 to the node $u$ corresponding to the intended receiver (or goal) and 0 to all other nodes. The GAT processes the graph with these extended node features $\mathbf{X}_u^* = [\mathbf{X}; \mathbf{e}_u] \in \mathbb{R}^{|\mathcal{V}| \times 26}$ and produces node-wise logits. A softmax is then applied over the logits of the defending players $v \in \mathcal{V}^-$ to yield a probability distribution representing their responsibilities $\left(\hat{y}_{u,v}\right)_{v \in \mathcal{V}^-}$ for defending the given action $v$:

$$\hat{\mathbf{y}}_u^- = \left(\hat{y}_{u,v}\right)_{v \in \mathcal{V}^-} = \mathrm{softmax}\left(\mathrm{MLP}_\phi(\mathbf{H}_u^-)\right), \tag{24}$$

where $\mathbf{H}_u^- = \left(\mathbf{h}_{u,v}\right)_{v \in \mathcal{V}^-}$ denotes the node embeddings of the defending players. The model is trained using only failed passes and blocked shots. For each instance, we construct the graph with the appropriate action-indicator feature, and train the model using cross-entropy against the identity of the defender who actually intercepted or blocked the action.

### 3.2. Unblocked-Shot Expected Goals Model

Because shots and goals occur infrequently during the match, the dataset used in this study does not provide enough samples to reliably train an Expected Goals (xG) [1] [15] model. However, estimating the xG of unblocked shots (c3) does not need to account for the possibility of the shot being blocked, and therefore does not depend on contextual information about surrounding players. This allows us to train the model using external event data with sufficient shot volume.

To this end, we utilize the Wyscout open event dataset [24], from which we extract shot features and labels from shots recorded in 1,941 matches. The dataset includes 45,945 shot attempts, where 34,987 were unblocked and 5,105 resulted in goals. We train a logistic regression model using six shot features: the $(x, y)$ location relative to the goal, the distance and angle to the goal, and binary indicators for whether the shot was taken from a set-piece and was a header.

## 4. Experiments

Since there is no ground truth measure of players' defensive ability, we evaluate the proposed DEFCON through several complementary approaches. First, Section 4.2 compares the predictive accuracy of the component GAT models against other baseline models. Also, Section 0 analyzes the correlation between players' aggregated defensive scores with their market values. Section 4.4 further provides qualitative validation by inspecting which players rank highest and comparing these rankings with market-value rankings. The datasets used in all experiments are described in Section 4.1.

### 4.1. Datasets

In this study, we utilize optical tracking data and event data collected from 564 matches of the 2023–24 and 2024–25 seasons of Dutch Eredivisie, provided by AFC Ajax. The tracking data contains player and ball positions sampled at 25Hz, while the event data consists of approximately 1,400 manually annotated on-ball actions per match. Since manually recorded event timestamps contain temporal inaccuracies, we synchronize all events with the tracking data using ELASTIC [17].

To examine how well the models generalize across different seasons, we use the 2023–24 season data to train all component models and evaluate them on the 2024–25 season. We consider passes, dribbles, and shots across both open play and set-piece situations for model training, with each model utilizing a different subset of actions. Note that although different subsets are used for training, all models are applied to every pass, dribble, and shot event in the test set, since EPV computation and defensive credit assignment requires component values for every possible attacking option. Table 2 describes the detailed statistics on the dataset.

**Table 2:** The number of actions used in each predictive task and the number of matches shared across all tasks. The term 'action' encompasses pass, dribble, and shot.

| Task | Action type | Training | Validation | Test | % Positive |
|---|---|---|---|---|---|
| (a1) Action selection | Action | 200,255 | 81,650 | 286,696 | – |
| (b1) Pass success | Pass & dribble | 194,680 | 79,433 | 279,328 | 76.80% |
| (c1) Goal-scoring | | | | | 1.59% |
| (c2) Goal-conceding | | | | | 0.63% |
| (b2) Shot blocking | Real shot | 5,575 | 2,217 | 7,368 | 26.20% |
| | Augmented shot | 3,773 | 1,552 | 0 | 100% |
| (d1) Responsibility | Failed action | 36,782 | 14,947 | 50,984 | – |
| # Matches | – | 200 | 81 | 283 | – |

## 4.2. Predictive Performance of Component Models

To evaluate the predictive performance of the component models used in DEFCON, we compare our GAT-based models with both GNN (GCN [18] and GIN [38]) and gradient-boosting (XGBoost [7] and CatBoost [26]) baselines. The applicable baselines differ by task due to differences in problem formulation. Pass success prediction (b1), shot-blocking prediction (b2), and outcome-conditioned goal-scoring (c1) and goal-conceding (c2) prediction are binary classification tasks, so they can be implemented using either GNNs or tabular boosting models. In contrast, action selection prediction (a1) and defender responsibility estimation (d1) require selecting one node among many, where boosting models cannot directly support. Therefore, we only compare GNN variants for these tasks.

For boosting models, we replace the graph input with tabular feature vectors. In pass-related tasks (b1, c1, c2), we construct input vectors by concatenating the the node features of the ball carrier and the intended receiver. For shot-blocking (b2), we use only the ball carrier's node features.

Table 3 reports the resulting performance in terms of F1-score, AUC, and Brier score for binary classification tasks (b1, b2, c1, c2), and accuracy, cross entropy (CE), and mean reciprocal rank (MRR) for multi-class classification tasks (a1, d1). Overall, GAT outperforms all GNN variants and is the strongest model for pass success prediction (b1) across all methods. Its superior performance reflects the power of attention-weighted message-passing, which enables the model to modulate the influence of surrounding players. Pass success prediction benefits from abundant training data and minimal class imbalance, allowing GAT to fully leverage relational information and outperform both GNN baselines and boosting models.

Interestingly, boosting models outperform GNNs on the shot-blocking prediction (b2) and goal prediction (c1, c2) tasks, for different potential reasons. For shot-blocking, GNN models treat the task as a graph classification problem, which requires aggregating node embeddings via average pooling. This inevitably discards some relational information and places GNNs at a structural disadvantage compared to boosting models. Meanwhile, the goal prediction tasks suffer from extreme class imbalance, making it difficult for GNNs to learn stable relational patterns.

Despite these limitations, GNNs achieve Brier scores comparable to boosting models across all tasks, indicating that they still produce well-calibrated probability estimates. Such calibration quality is important, since our objective is not simply to classify outcomes, but to reliably estimates underlying probabilities for accurate EPV computation and credit assignment. Given these considerations, and to maintain architectural consistency across all component models, we adopt GAT as the unified model architecture in all subsequent experiments and applications.

**Table 3:** Performance of component model variants.

| Task | Metric | XGBoost | CatBoost | GCN | GIN | GAT |
|---|---|---|---|---|---|---|
| (a1, multi-class) Action selection | Accuracy | – | – | 0.6080 | 0.6119 | **0.6738** |
| | CE | – | – | 1.1003 | 1.0925 | **0.9239** |
| | MRR | – | – | 0.7555 | 0.7581 | **0.7996** |
| (b1, binary) Pass success | F1 | 0.9059 | 0.9060 | 0.9104 | 0.9105 | **0.9115** |
| | AUC | 0.9124 | 0.9115 | 0.9149 | 0.9151 | **0.9167** |
| | Brier | 0.0980 | 0.0984 | 0.0942 | 0.0941 | **0.0933** |
| (b2, binary) Shot blocking | F1 | 0.4485 | **0.4491** | 0.4232 | 0.4117 | 0.3984 |
| | AUC | 0.6845 | **0.6896** | 0.6373 | 0.6402 | 0.6717 |
| | Brier | 0.2070 | **0.2014** | 0.2313 | 0.2268 | 0.2037 |
| (c1, binary) Goal-scoring | F1 | 0.0884 | **0.0951** | 0.0780 | 0.0721 | 0.0759 |
| | AUC | 0.6873 | **0.7056** | 0.6823 | 0.6818 | 0.6834 |
| | Brier | 0.0149 | 0.0149 | **0.0148** | **0.0148** | **0.0148** |
| (c2, binary) Goal-conceding | F1 | **0.0070** | 0.0000 | 0.0000 | 0.0000 | 0.0000 |
| | AUC | 0.6030 | **0.6478** | 0.6168 | 0.6174 | 0.6172 |
| | Brier | **0.0031** | **0.0031** | 0.0032 | 0.0032 | 0.0032 |
| (d1, multi-class) Responsibility | Accuracy | – | – | 0.2439 | 0.2361 | **0.5021** |
| | CE | – | – | 2.3282 | 2.3399 | **1.4036** |
| | MRR | – | – | 0.4868 | 0.4792 | **0.6940** |

## 4.3. Correlation with Market Values

To check how well the proposed defensive credit reflects players' actual defensive ability, we examine its correlation with players' market values. Specifically, we consider 261 outfield players from Eredivisie 2024–25 who played at least 900 minutes and compute each player's average defensive credit per 90 minutes. Market values are collected from Transfermarkt[4] at the end of the season and transformed into log scale to mitigate the heavy-tailed distribution, where a small number of star players exhibit disproportionately higher valuations.

We analyze correlations for three types of defensive scores:

(a) Intercept: positive credits earned through on-ball defensive actions such as intercepting passes (including tackling against dribbles treated as self-passes) or blocking shots,
(b) Concede: penalties incurred from conceding threatening attacks or committing fouls, and
(c) Net credit: the total credit across all types of defense.

To better understand positional differences, we report correlations both across all players and within major positional groups: defenders (DF), midfielders (MF), and forwards (FW), with the corresponding scatterplots shown in Figure 8. We further analyze three defensively oriented sub-positions: center-backs (CB), side-backs (SB), and central/defensive midfielders (CM), as visualized in Figure 9. Table 4 summarizes the Pearson correlation coefficients against the log-scale market values against these groups.

[4] https://www.transfermarkt.com

**Table 4:** Pearson correlation coefficients between defensive credits and log-scale market values for regular players in Eredivisie 2024–25. Coefficients greater than 0.5 are highlighted in bold.

| Role | Players | Intercept | | | Concede | | | | Net credit |
|---|---|---|---|---|---|---|---|---|---|
| | | Pass | Shot | Total | Pass | Shot | Foul | Total | |
| Total | 261 | −0.159 | −0.242 | −0.204 | 0.442 | 0.217 | 0.098 | 0.344 | 0.383 |
| DF | 99 | −0.220 | −0.317 | −0.317 | **0.665** | 0.272 | 0.133 | **0.545** | 0.451 |
| MF | 132 | −0.131 | −0.175 | −0.157 | **0.605** | 0.169 | 0.008 | 0.425 | 0.346 |
| FW | 30 | −0.329 | −0.230 | −0.193 | 0.481 | 0.074 | 0.087 | 0.333 | 0.229 |
| CB | 46 | −0.497 | −0.447 | −0.596 | **0.752** | **0.643** | 0.059 | **0.754** | **0.563** |
| SB | 53 | −0.110 | −0.281 | −0.235 | **0.600** | 0.123 | 0.214 | **0.518** | 0.360 |
| CM | 66 | −0.311 | −0.287 | −0.344 | **0.755** | 0.274 | 0.158 | **0.621** | 0.337 |

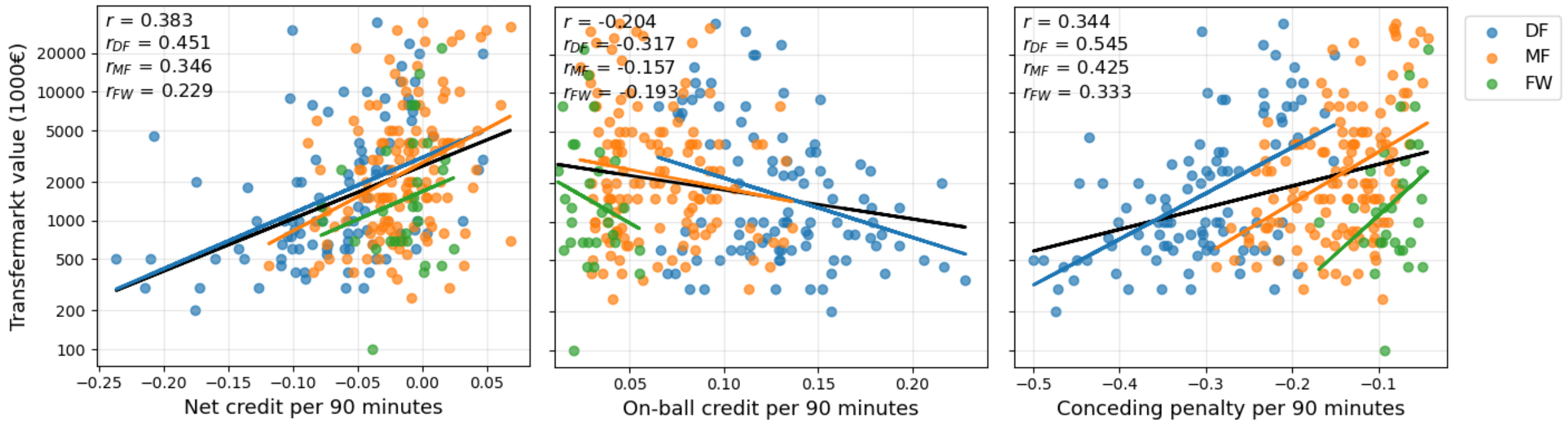

**Figure 8:** Relationships between category-wise defensive credits and log-scale market values for all regular players in Eredivisie 2024–25, with points and trend lines color-coded by major positional group.

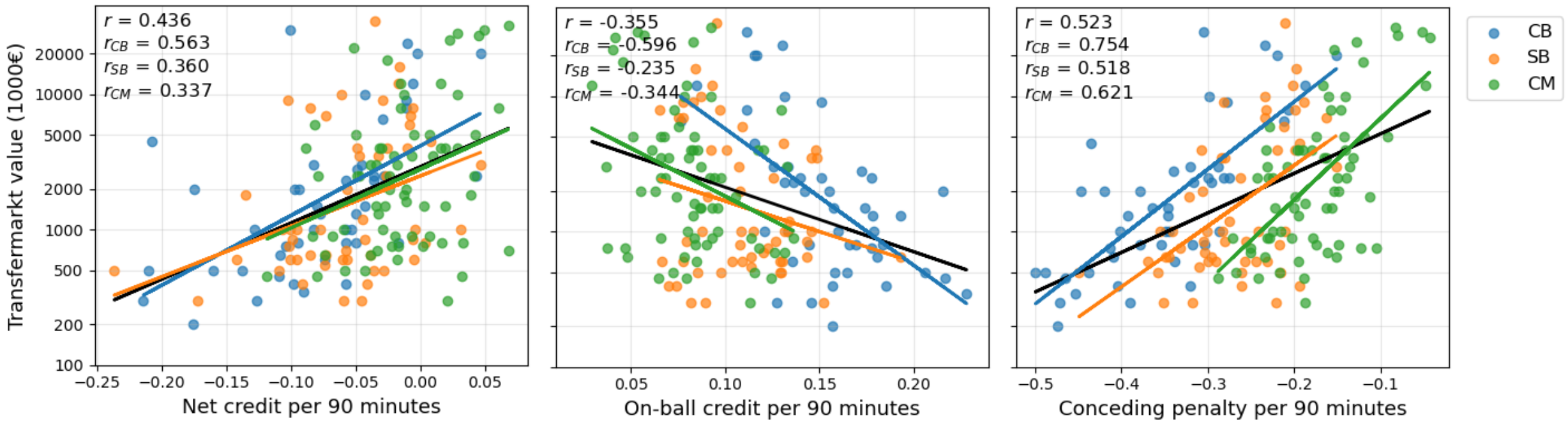

**Figure 9:** Relationships between category-wise defensive credits and log-scale market values for defensively orientied positions in Eredivisie 2024–25, with points and trend lines color-coded by positional subgroup.

Across all players, net defensive credit exhibits a clear positive correlation with market value, and this relationship becomes stronger for more defensively oriented roles. In particular, center-backs show a strong correlation of 0.563, indicating that our metric effectively captures aspects of defensive quality reflected in the real-world market.

In contrast, credits obtained from on-ball defensive actions display a negative correlation with market value. This counter-intuitive trend reflects a fundamental limitation of existing action-based defensive evaluation: players on weaker teams naturally face more defensive situations and therefore become overestimated by accumulating more on-ball defensive actions. This underscores the value of our approach, which rewards not only direct defensive actions but also the prevention of conceding threatening attacks.

Notably, penalties for conceding threatening attacks show the strongest alignment with market value among all components. Across every positional group, players with smaller absolute penalties (i.e., those who allow fewer dangerous attacks) tend to have higher market valuations. Correlations for all defensive roles (DF, CB, SB, and CM) exceed 0.5, with center-backs showing a particularly high correlation of 0.754. Interestingly, pass-conceding penalties are highly informative across all positions, whereas shot-conceding penalties show a strong correlation only for center-backs. This aligns with common intuition that center-backs bear primary responsibility for defending shots, while defending passes is a shared duty across positions. Overall, these results demonstrate that preventing opponents' dangerous actions is far more important than performing many on-ball defensive actions (summarized as "better prevent than tackle").

## 4.4. Qualitative Validation Through Player Rankings

Beyond the correlation analysis, we also conduct a qualitative inspection of which players appear at the top of each defensive metric. Tables 5–7 list the top-10 center-backs according to net credit, defensive action ("intercept") credit, and conceding penalty, respectively. The rightmost column reports each player's Transfermarkt market value, with the number in parentheses indicating their rank among the 46 center-backs who played at least 900 minutes in the 2024–25 season.

The results are consistent with our earlier findings. For net credit (Table 5), 7 of the top-10 players are also ranked within the top-10 by market value, suggesting strong alignment between our metric and practitioners' valuation. A similar pattern appears for conceding penalties (Table 7), where high-value players tend to cluster near the top. In contrast, players ranked highly by defensive action credit (Table 6) generally have low market values. They accumulated many interceptions or tackles because their teams were exposed to more dangerous situations, emphasizing that values obtained from observed actions alone do not reliably represent individual defensive ability.

While market value is influenced by many confounding factors such as age, nationality, and contract length, it nevertheless correlates with overall perceived ability, making it a useful indirect validation signal. In practical settings, clubs can leverage DEFCON's metrics to automatically filter players who exceed certain thresholds in desired defense categories, substantially narrowing scouting pools. Instead of manually reviewing large numbers of players, analysts can focus their time on a targeted shortlist identified through our trustworthy defensive valuation.

**Table 5:** Top-10 center-backs ranked by net defensive credit.

| Rank | Team | Player | Intercept | Concede | Net credit | Market value |
|---|---|---|---|---|---|---|
| 1 | PSV Eindhoven | O. Boscagli | 0.1165 | –0.1511 | 0.0464 | €20.0m (3) |
| 2 | Sparta Rotterdam | M. Young | 0.1778 | –0.2756 | 0.0434 | €2.5m (14) |
| 3 | PSV Eindhoven | R. Flamingo | 0.1153 | –0.2200 | –0.0027 | €20.0m (3) |
| 4 | Ajax Amsterdam | Y. Baas | 0.0848 | –0.1872 | –0.0056 | €12.0m (5) |
| 5 | Ajax Amsterdam | J. Šutalo | 0.1304 | –0.2335 | –0.0098 | €24.0m (2) |
| 6 | Feyenoord Rotterdam | T. Beelen | 0.1114 | –0.2063 | –0.0110 | €8.0m (8) |
| 7 | AZ Alkmaar | Alexandre Penetra | 0.1507 | –0.2774 | –0.0110 | €9.0m (7) |
| 8 | FC Twente | A. Van Hoorenbeeck | 0.1838 | –0.3184 | –0.0174 | €0.8m (30) |
| 9 | FC Twente | M. Bruns | 0.1399 | –0.2818 | –0.0278 | €2.5m (14) |
| 10 | FC Twente | M. Hilgers | 0.0762 | –0.2127 | –0.0290 | €6.5m (9) |

**Table 6:** Top-10 center-backs ranked by "intercept" credit (i.e., highest credits from defensive actions).

| Rank | Team | Player | Intercept | Concede | Net credit | Market value |
|---|---|---|---|---|---|---|
| 1 | Almare City FC | J. Jacobs | 0.2281 | −0.4539 | −0.0907 | €350k (43) |
| 2 | RKC Waalijk | R. Van Eijma | 0.2165 | −0.4652 | −0.1090 | €450k (40) |
| 3 | NAC Breda | L. Greiml | 0.2155 | −0.4469 | −0.0970 | €2,000k (19) |
| 4 | RKC Waalijk | D. Van den Buijs | 0.2059 | −0.4995 | −0.1598 | €500k (37) |
| 5 | Heracles Almelo | I. Mesík | 0.1932 | −0.3921 | −0.0770 | €1,300k (24) |
| 6 | Go Ahead Eagles | G. Nauber | 0.1855 | −0.4042 | −0.0978 | €400k (41) |
| 7 | FC Twente | A. Van Hoorenbeeck | 0.1838 | −0.3488 | −0.0174 | €800k (30) |
| 8 | Heracles Almelo | D. Mirani | 0.1807 | −0.3462 | −0.1081 | €650k (35) |
| 9 | SC Heerenveen | S. Kersten | 0.1787 | −0.2756 | −0.0579 | €1,000k (26) |
| 10 | Fortuna Sittard | S. Adewoye | 0.1781 | −0.4138 | −0.0500 | €1,300k (24) |

**Table 7:** Top-10 center-backs ranked by "concede" credit (i.e., lowest penalties for allowing attacks).

| Rank | Team | Player | Intercept | Concede | Net credit | Market value |
|---|---|---|---|---|---|---|
| 1 | PSV Eindhoven | O. Boscagli | 0.1165 | −0.1511 | 0.0464 | €20.0m (3) |
| 2 | Ajax Amsterdam | Y. Baas | 0.0848 | −0.1872 | −0.0056 | €12.0m (5) |
| 3 | Feyenoord Rotterdam | T. Beelen | 0.1114 | −0.2063 | −0.0110 | €8.0m (8) |
| 4 | FC Twente | M. Hilgers | 0.0762 | −0.2127 | −0.0290 | €6.5m (9) |
| 5 | PSV Eindhoven | R. Flamingo | 0.1153 | −0.2200 | −0.0027 | €20.0m (3) |
| 6 | Ajax Amsterdam | J. Šutalo | 0.1304 | −0.2335 | −0.0098 | €24.0m (2) |
| 7 | Sparta Rotterdam | M. Young | 0.1778 | −0.2756 | 0.0434 | €2.5m (14) |
| 8 | AZ Alkmaar | Alexandre Penetra | 0.1507 | −0.2774 | −0.0110 | €9.0m (7) |
| 9 | FC Twente | M. Bruns | 0.1399 | −0.2818 | −0.0278 | €2.5m (14) |
| 10 | FC Utrecht | M. Van der Hoorn | 0.1206 | −0.2875 | −0.0529 | €1.0m (26) |

# 5. Practical Applications

To illustrate how DEFCON can support real-world analysis in professional clubs, this section showcases several practical applications that leverage its defensive valuations across temporal (Section 5.1), spatial (Section 5.2), and relational (Section 5.3) dimensions.

## 5.1. Interactive Timeline

To support practical match-analysis workflows, we develop an interactive timeline visualization (Figure 10) that enables analysts to efficiently retrieve key defensive moments in terms of credit gain and loss. The tool is implemented using Plotly[5], allowing users to zoom into specific time intervals and inspect events by hovering over the timeline, which reveals the action index, action type, involved players, and assigned defensive credit.

This interface allows users to quickly spot when and how each player performed good defense or allowed dangerous attacks throughout the match, simply by scanning for spikes in credit, rather than manually reviewing the entire match. After finding an important moment, analysts can identify the corresponding timestamp or the action index and then use them to visualize or replay the associated scene. In the future, we plan to integrate this prototype with common visualization tools such as Power BI[6] or video analysis tools like Hudl Sportscode[7] to enable seamless navigation from a colored event in the timeline to the exact video clip. Such integration can significantly streamline and accelerate data-driven post-match analysis in professional clubs.

[5] https://plotly.com

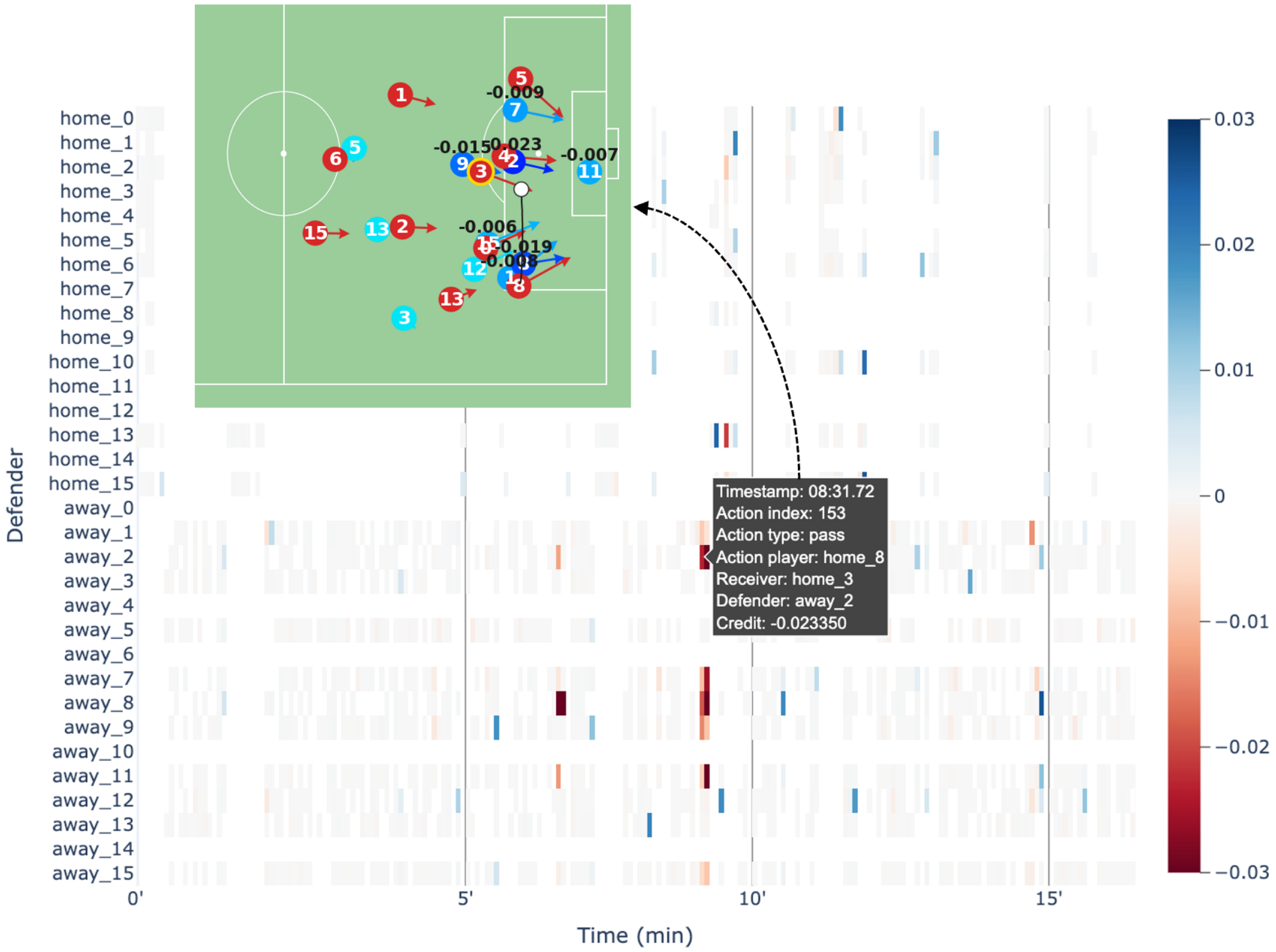

**Figure 10:** Retrieval of key defensive moments using an interactive timeline for player-wise credit gain and loss, where blue indicates positive credit and red indicates penalties.

## 5.2. Spatial Tendency Across Pitch Zones

To understand where on the pitch teams gain and lose defensive value, we visualize the spatial distribution of defensive credits in the form of heatmaps overlaid on the pitch, as shown in Figure 11. These visualizations reveal spatial patterns of each team's offensive and defensive performance that cannot be captured through action counts alone.

For the away team in the figure, although they accumulated a large amount of positive credit through on-ball defensive actions and disturbances, these gains were concentrated in front of their own goal. Crucially, they also incurred much larger penalties in these same high-risk zones, indicating repeated exposure to dangerous attacks. This pattern reflects their reactive defensive performance: forced deep into their box, often succeeding in last-moment interventions but still allowing the opponent to reach threatening areas too frequently.

In contrast, the home team exhibited a very different defensive profile. Because they were rarely pushed deep into their own box, their total positive credit near goal was smaller. However, the credits they did earn were spread across the entire pitch, most notably the central regions inside the opponent's half, which appear as a prominent blue area in the Home-Disturb heatmap. Their high "disturb" value in this zone suggests effective high pressing that reduced the success probability of the opponent's build-up before dangerous situations could emerge.

[6] https://www.microsoft.com/en-us/power-platform/products/power-bi
[7] https://www.hudl.com/en_gb/products/sportscode

There is also a notable contrast in both teams' half-spaces, i.e., the vertical lanes located between the central lane and the flank. In these regions, the home team conceded almost no penalty, while the away team incurred a large amount. Although the away team got relatively low penalties on the flanks, indicating that they allowed fewer wide attacks, they were repeatedly broken down through the more dangerous central and half-space channels around the box.

Together, these spatial patterns show how DEFCON provides interpretable insights into team defensive behavior, enabling analysts to identify intense battlefields on the pitch, diagnose teams' structural weaknesses, and evaluate their pressing effectiveness.

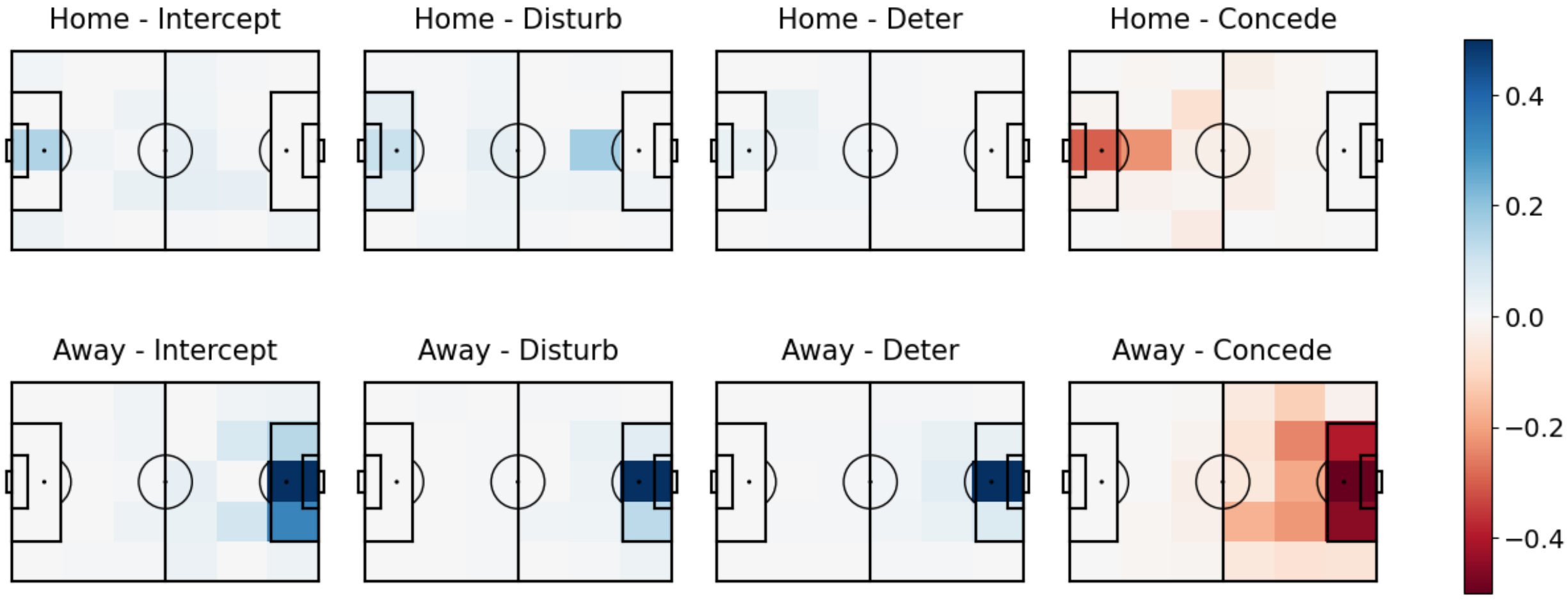

**Figure 11:** Spatial distribution of defensive credits across pitch zones for the match illustrated in Figure 6. The top and bottom rows display the credit heatmaps for the home and the away teams, respectively, with values separated into the four defense categories defined in Section 2.6.

## 5.3. Pairwise Attacker-Defender Analysis

Unlike existing offense-centered evaluation frameworks which do not identify defensive counterparts of each attacking action, our approach explicitly determines the defending responsibilities associated with every on-ball attacking action. Thus, it enables pairwise attacker-defender analysis about how much defensive credit each defender gained or lost against each opposing attacker, as visualized in Figure 12. This allows analysts to identify which defenders held superiority or inferiority over specific opponents.

Moreover, combining these pairwise values with spatial context such as players' roles or average on-field locations provides further insights into how and where these credits or penalties emerged. For example, the lower left matrix in Figure 12 shows that many away defenders earned large credit against home 4, who appears as a central forward in Figure 13. When interpreted alongside the Away-Intercept and Away-Disturb heatmaps in Figure 11, which highlight intense defensive activity around their own penalty box, this pattern implies away defenders' frequent last-moment interventions against home 4's scoring attempts in dangerous situations.

In contrast, the lower-right matrix indicates that the away team's penalties were distributed across a wide range of home attackers, rather than concentrated on a single player. For instance, Figure 13 shows that during the first half, the left fullback away 1 conceded significant value to home 8 (right winger), as well as home 0 (right fullback) and several opposing midfielders. Meanwhile, the center back away 2 effectively handled home 4 (center forward), incurring relatively small penalties against him. However, away 2 instead received huge penalties from home 3 (central midfielder), indicating a vulnerability in defending threatening passes made from the opposing midfield.

Overall, this pairwise attacker-defender analysis provides actionable insights into matchup-specific strengths and weaknesses. Such information can support various real-world decision-making processes, such as selecting lineups that best exploit or mitigate specific matchups and scouting players whose defensive style fits the tactical needs of the team.

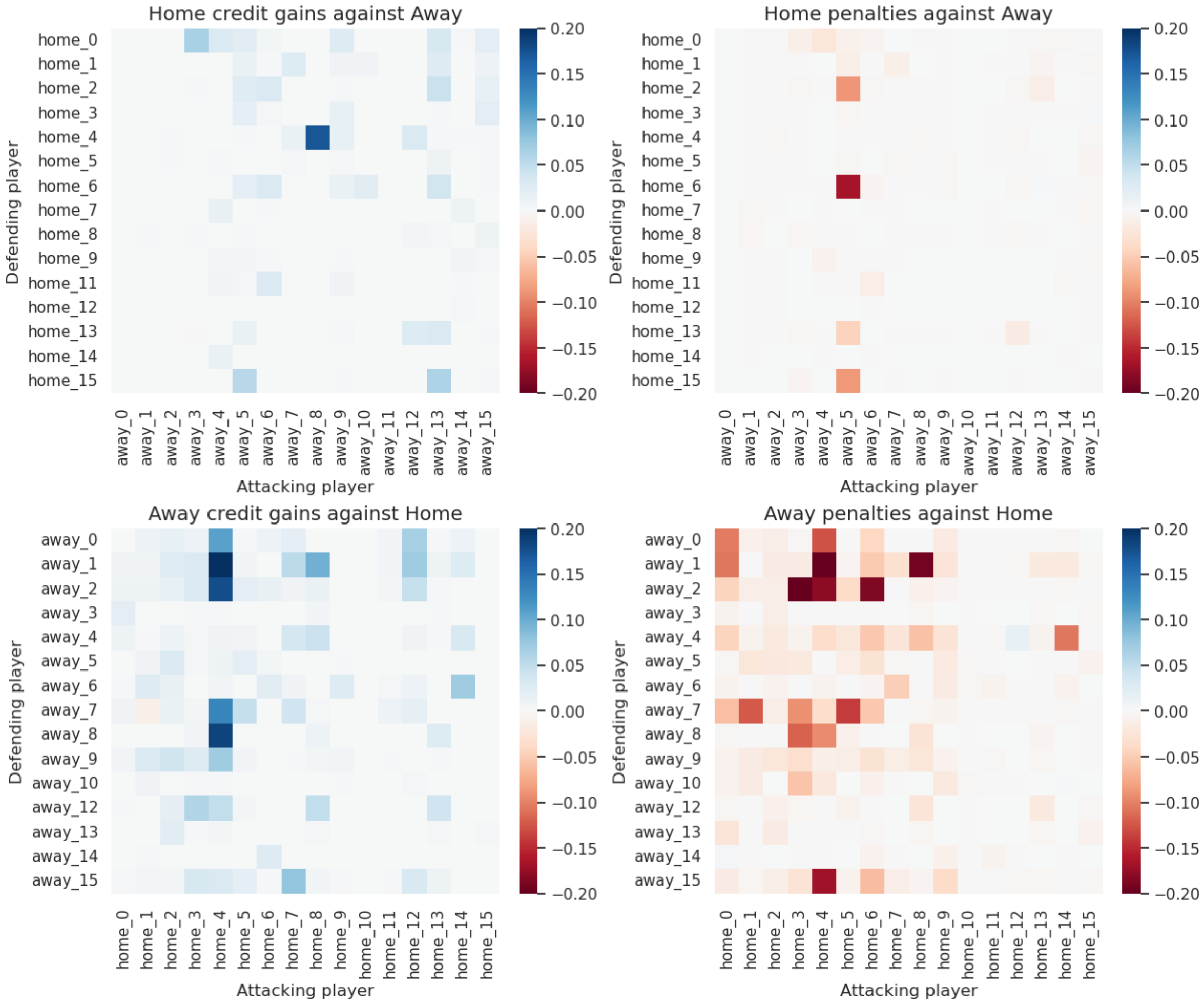

**Figure 12:** Pairwise defensive credit matrices, visualizing how much defensive credit (left) or penalty (right) each defending player accumulated against each opposing attacker in the match.

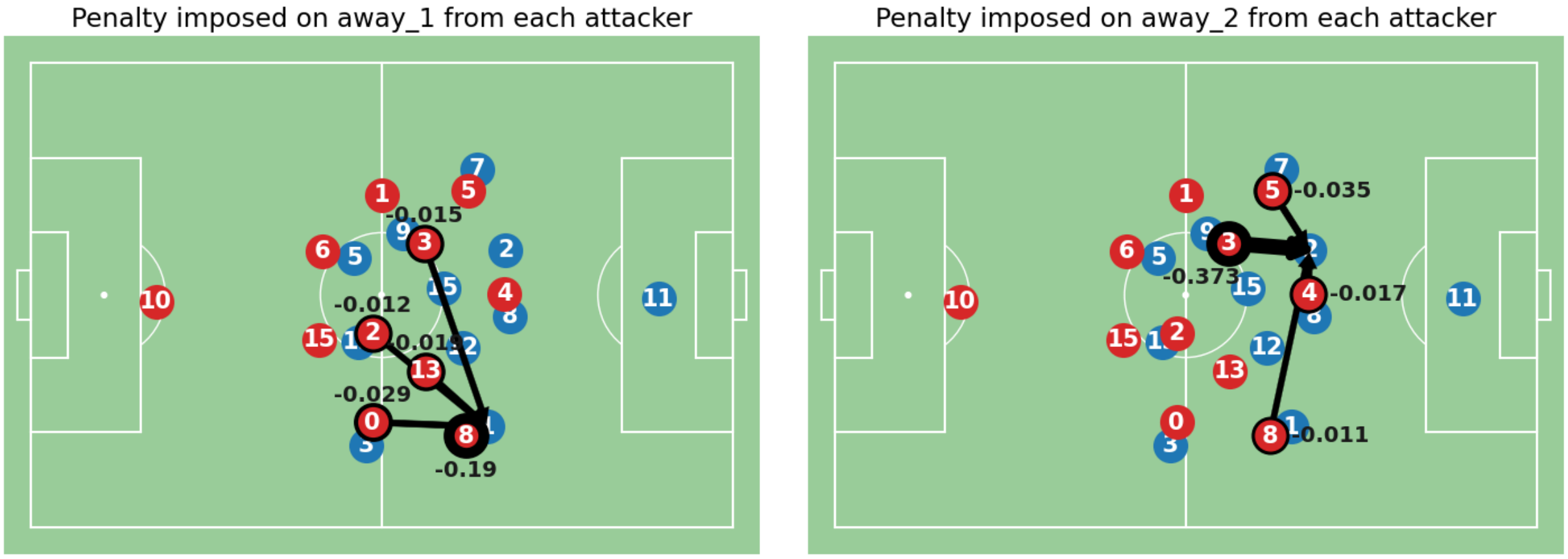

**Figure 13:** Visualziation of pairwise penalties imposed on away defenders against home attackers during the first half of the match. Player locations correspond to their average locations in the half, and the thickness of each arrow is proportional to the magnitude of penalty from the attacker to the defender of interest.

# 6. Conclusions

This paper proposes DEFCON, a comprehensive framework for fine-grained quantification of defensive contributions in soccer. While most existing data-driven approaches have largely focused on on-ball actions, DEFCON addresses the longstanding challenge of also evaluating what they prevent or concede and how responsible they are for defending each attacking option.

For every attacking on-ball action, DEFCON classifies the situation based on pass success or failure, EPV increase or decrease, and whether a shot is blocked, and assigns defensive credit to individual players according to principled, scenario-specific rules. To enable these computations, it employs Graph Attention Networks to estimate key component values such as action success probabilities, success-conditioned EPVs, and defender responsibilities with high contextual fidelity. The resulting defensive credits align well with players' actual defensive ability, revealing that the penalty for conceding dangerous attacks exhibits a far stronger correlation with market value than the credit for on-ball defensive actions alone. Finally, we present practical applications of DEFCON in temporal, spatial, and relational analyses, providing practitioners with actionable, interpretable insights with multiple levels of resolution.

Looking ahead, several extensions of DEFCON present promising avenues for future work. One direction is to integrate our defensive valuation with established offensive metrics such as xT [32], VAEP [8], and GIM [20], enabling holistic assessment of players' offensive and defensive contributions. Since a player's tactical roles often change across matches and even within a single match, another direction is to incorporate algorithms for detecting time-varying formations and player roles [4] [5] [16] to separate the player's defensive performance by role and compare multiple players taking the same role. Finally, we aim to extend our applications to tactical analysis, such as identifying which formation best defend defending against specific opponent structures, or diagnosing which role become vulnerable when deploying a particular lineup against them.